\documentclass[final,5p,times,twocolumn]{elsarticle}
\makeatletter
\def\ps@pprintTitle{%
 \let\@oddhead\@empty
 \let\@evenhead\@empty
 \let\@oddfoot\@empty
 \let\@evenfoot\@empty}
\makeatother
\usepackage{graphicx}
\usepackage{multirow}
\usepackage{times}
\usepackage{amsmath}
\usepackage{amssymb}
\usepackage{array}
\usepackage{color}
\usepackage{subcaption}
\usepackage{mathtools}
\usepackage{url}
\usepackage{bm}
\usepackage{breqn}

\journal{Neurocomputing}

\usepackage{fancyhdr}
\begin{document}
\begin{frontmatter}

\title{Activation Functions in Deep Learning: A Comprehensive Survey and Benchmark}

\author{Shiv Ram Dubey$^1$, Satish Kumar Singh$^1$, Bidyut Baran Chaudhuri$^2$}
\address{
$^1$Computer Vision and Biometrics Laboratory, Indian Institute of Information Technology, Allahabad, India. \\
$^2$Techno India University, Kolkata, India and Indian Statistical Institute, Kolkata, India.
\\[\bigskipamount]
{\texttt{srdubey@iiita.ac.in, sk.singh@iiita.ac.in, bidyutbaranchaudhuri@gmail.com}
\\[\bigskipamount]
\textcolor{red}{This paper is accepted in Neurocomputing. Copyright will be transferred to Elsevier.}
}}

\begin{abstract}
Neural networks have shown tremendous growth in recent years to solve numerous problems. Various types of neural networks have been introduced to deal with different types of problems. However, the main goal of any neural network is to transform the non-linearly separable input data into more linearly separable abstract features using a hierarchy of layers. These layers are combinations of linear and nonlinear functions. The most popular and common non-linearity layers are activation functions (AFs), such as Logistic Sigmoid, Tanh, ReLU, ELU, Swish and Mish. In this paper, a comprehensive overview and survey is presented for AFs in neural networks for deep learning. Different classes of AFs such as Logistic Sigmoid and Tanh based, ReLU based, ELU based, and Learning based are covered. Several characteristics of AFs such as output range, monotonicity, and smoothness are also pointed out. A performance comparison is also performed among 18 state-of-the-art AFs with different networks on different types of data. The insights of AFs are presented to benefit the researchers for doing further research and practitioners to select among different choices. The code used for experimental comparison is released at: \url{https://github.com/shivram1987/ActivationFunctions}.
\end{abstract}
\end{frontmatter}

% \keywords{Activation Functions, Neural Networks, Convolutional Neural Networks, Deep Learning, Overview, Recurrent Neural Networks}

% \maketitle

\section{Introduction}
\label{Introduction}
In recent years, deep learning has shown a tremondous growth to solve the challenging problems such as 
% facial analysis \cite{agbo2021deep}, \cite{punyani2020neural}, 
% prediction rating \cite{khan2021deep}, 
% sentiment analysis \cite{wadawadagi2020sentiment}, \cite{yadav2020sentiment}, 
% hyperspectral image analysis \cite{wang2021review}, 
% image synthesis and semantic manipulations \cite{abdolahnejad2020deep}, 
% digital images augmentation \cite{khalifa2021comprehensive}, 
object detection \cite{shao2022deep}, semantic segmentation \cite{mo2022review}, person re-identification \cite{guo2022jac},
image retrieval \cite{dubey2021decade}, anomaly detection \cite{xia2022gan}, skin disease diagnosis \cite{li2021skin}, and many more.
Various types of neural networks have been defined in deep learning to learn abstract features from data, such as Multilayer Perceptron (MLP) 
\cite{dagli2012artificial}, Convolutional Neural Networks (CNN) \cite{alexnet}, 
% \cite{khan2020survey}, 
Recurrent Neural Networks (RNN) \cite{graves2013speech}, and Generative Adversarial Networks (GAN) \cite{babu2020pcsgan}. 
The important aspects of neural networks include 
weight initialization \cite{liu2022weight},
% \cite{narkhede2021review}, 
loss functions \cite{srivastava2019performance},
different layers \cite{basha2020impact},
overfitting \cite{xu2019overfitting},
% regularization \cite{moradi2020survey},
% ovefitting control \cite{bejani2021systematic}, 
% activation functions \cite{tan2014comparative}, 
and optimization \cite{dubey2019diffgrad}.

The activation functions (AFs) play a very crucial role in neural networks \cite{duch1999survey} by learning the abstract features through non-linear transformations. 
Some common properties of the AFs are as follows: a) it should add the non-linear curvature in the optimization landscape to improve the training convergence of the network; b) it should not increase the computational complexity of the model extensively; c) it should not hamper the gradient flow during training; d) it should retain the distribution of data to facilitate the better training of the network.
Several AFs have been explored in recent years for deep learning to achieve the above mentioned properties.
This survey is dedicated to the developments in the area of AFs in neural networks. The insights of the different AFs are presented along with the reasoning to benefit the deep learning community. 
The major contributions of this survey are outlined as follows:

\begin{enumerate}
    \item This survey provides a detailed classification for a wide range of AFs. It also includes the AFs very comprehensively, including Logistic Sigmoid/Tanh, Rectified Unit, Exponential Unit, and Adaptive AFs.
    \item This survey enriches the reader with the state-of-the-art AFs with analysis from various perspectives. It specifically covers the progress in AFs for deep learning.
    \item This survey also summarizes the AFs with brief highlights and important discussions to depict its suitability for different types of data (Refer to Table \ref{tab:act_summary}).  
    \item This survey is compared with the existing survey and performance analysis to show its importance (Refer to Table \ref{tab:existing_survey}).
    \item This paper also presents the performance comparisons on 4 benchmark datasets of different modalities using 18 state-of-the-art AFs with different types of networks (Refer to Tables \ref{tab:results_cifar}, \ref{tab:results_cifar100} and \ref{tab:results_language_speech}).
\end{enumerate}

% An overview of neural networks is presented in Section \ref{nn}. 
The evolution of AFs is illustrated in Section \ref{evolution}. The progress in Logistic Sigmoid and Tanh, rectified, exponential, adaptive and miscellaneous AFs are summarized in Section \ref{sigmoid}, \ref{relu}, \ref{elu}, \ref{learning}, and \ref{other}, respectively. Some aspects of AFs are discussed in Section \ref{aspects}. A comprehensive performance analysis is conducted in Section \ref{performance}. A summary with conclusions and recommendations is provided in Section \ref{conclusion}.

\section{Evolution of Activation Functions}
\label{evolution}
\begin{figure}[!t]
\centering
\includegraphics[width=\columnwidth]{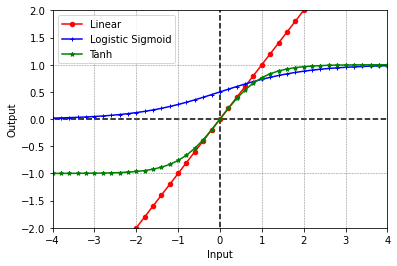}%<left> <lower> <right> <upper>
\caption{An illustration of Linear, Logistic Sigmoid and Tanh AFs.}
\label{fig:act}
\end{figure}

\begin{figure}[!t]
\centering
\includegraphics[trim={2.5cm 4.5cm 2.7cm 0.5cm},clip, width=\columnwidth]{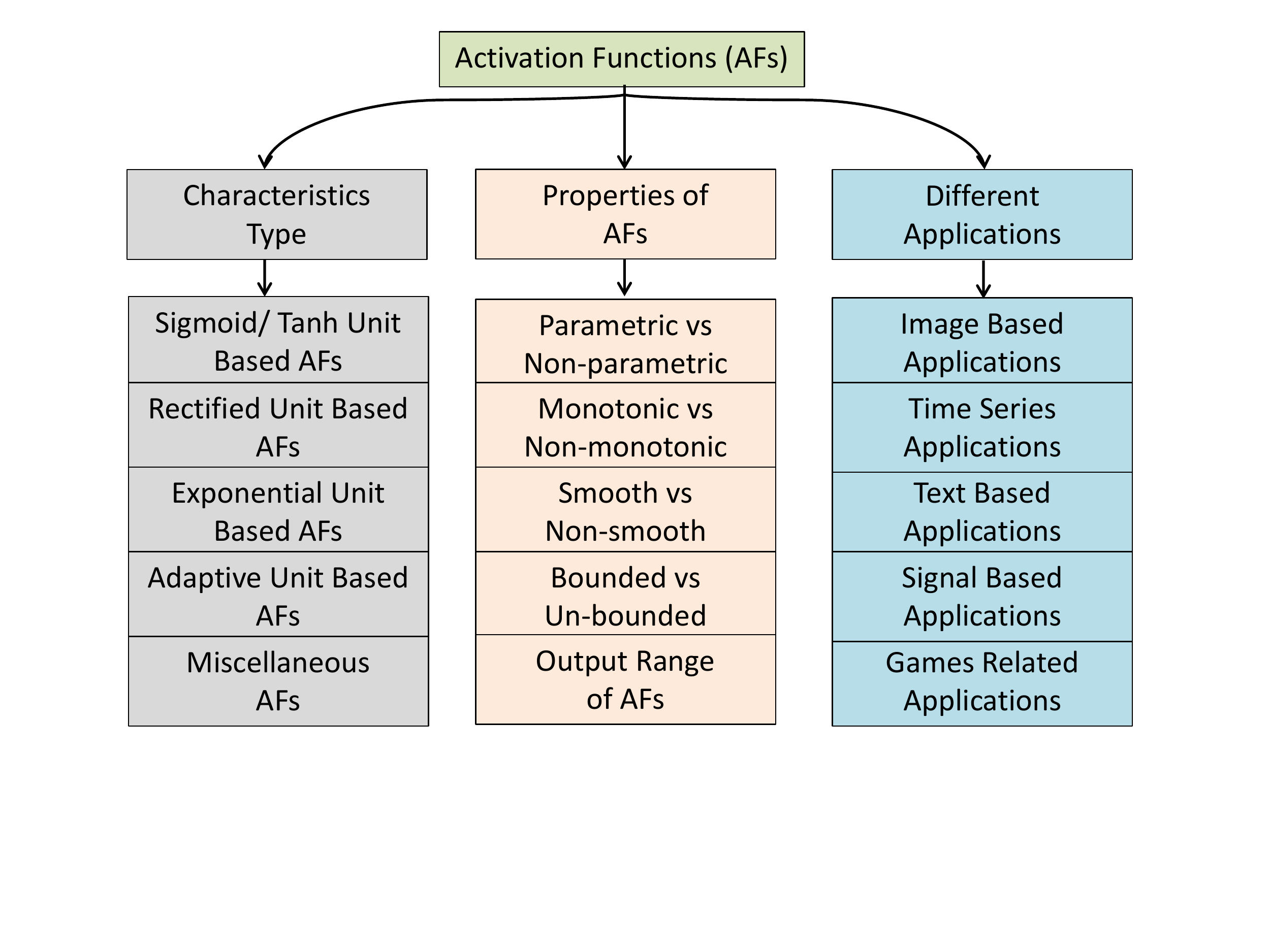}%<left> <lower> <right> <upper>
\caption{Classification of activation functions.}
\label{fig:af}
\end{figure}

% In the initial days, the neural networks used the Step function as the AF inspired from the biological neurons firing \cite{rolls1998neural}. The Step function outputs $1$ if the input value is greater than a threshold, otherwise $0$. In practice, the Step AF based neural network is hard to optimize as the step function is not differentiable at zero and the gradient for positive and negative input is zero.
A linear function can be thought of as a simple AF which outputs $c\times x$ for input $x$ with $c$ as a constant. 
The linear AF is illustrated in Fig. \ref{fig:act} for $c = 1$, i.e., identity function.
Note that the linear AF does not add non-linearity into the network. However, the non-linearity needs to be introduced in the neural networks. Otherwise, a neural network produces the output as a linear function of inputs inspite of having several layers. Moreover, in practice data is generally not linearly separable; hence, the non-linear layers help to project the data in non-linear fashion in feature space which can be used with different objective functions.
This section provides an overview of the evolution of AFs for deep learning. 
A classification is presented in Fig. \ref{fig:af} in terms of the different properties and characteristic types. 

% \subsection{Logistic Sigmoid/Tanh Unit Based Activation Functions}
\textit{Logistic Sigmoid/Tanh Unit Based Activation Functions:}
In order to introduce the non-linearity into the neural networks, the Logistic Sigmoid and Tanh AFs have been used in the early days. The firing of bilogical neurons was the motivation of using the Logistic Sigmoid and Tanh AFs with artificial neurons.
The Logistic Sigmoid AF is a very popular and traditional non-linear function. It is given as, 
\begin{equation}
\text{Logistic Sigmoid}(x) = \frac{1}{1+e^{-x}}.
\end{equation}
This AF squashes the output between [$0$, $1$] as shown in Fig. \ref{fig:act}. 
The output of the Logistic Sigmoid function is saturated for higher and lower inputs, which leads to vanishing gradient problem. The vanishing gradient problem depicts to a scenario where the gradient of objective function w.r.t. a parameter becomes very close to zero and leads to almost no update in the parameters during the training of the network using stochastic gradient descent technique. Hence, the training is almost killed under vanishing gradient scenario. Moreover, the output not following a zero-centric nature leads to poor convergence. 
% The Logistic Sigmoid function is computationally expensive due to exponential operation. 
The Tanh function has also been used as the AF in neural networks. It is similar to the Logistic Sigmoid function while exhibiting the zero centric property as depicted in Fig. \ref{fig:act}. The Tanh function is written as,
\begin{equation}
\text{Tanh}(x) = \frac{e^{x}-e^{-x}}{e^{x}+e^{-x}}.
\end{equation}
The Tanh function also squashes the inputs, but in $[-1,1]$. The drawbacks of Logistic Sigmoid function such as vanishing gradient and computational complexity also exist with Tanh function. 
The Logistic Sigmoid and Tanh AFs majorly suffer from vanishing gradient. Several improvements have been proposed based on the Logistic Sigmoid and Tanh AFs which are described in Section \ref{sigmoid} in detail.

% \subsection{Rectified Linear Unit Based Activation Functions}
\textit{Rectified Linear Unit Based Activation Functions:}
The saturated output and increased complexity are the key limitations of above-mentioned Logistic Sigmoid and Tanh based AFs. The Rectified Linear Unit (ReLU) \cite{nair2010rectified} has become the state-of-the-art AF due to its simplicity and improved performance.
The ReLU was also used in the AlexNet model \cite{alexnet}. Various variants of ReLU have been investigated by tackling its drawbacks, such as non-utilization of negative values, limited non-linearity and unbounded output, as detailed in Section \ref{relu}.

\begin{table*}[!t]
    \caption{Advantage and disadvantage of primary AFs.}
    \centering
    {\footnotesize
    \begin{tabular}{cccccc}
    \hline
    \multirow{2}{*}{AFs} & Diminishing & Limited &  Optimization & Lack of & Computational \\
    & gradients & non-linearity &  difficulty & adaptibility & inefficiency \\
    \hline
    Sigmoid & Yes & No & Yes & Yes & Yes \\
    Tanh & Yes & No & Partial & Yes & Yes \\
    ReLU & Partial & Yes & Partial & Yes & No \\
    ELU & No & Partial & No & Yes & Partial \\
    APL & No & Partial & No & No & No \\
    Swish & No & Partial & No & No & Partial \\
    \hline
    \end{tabular}
    }
    \label{tab:evolution_comparison}
\end{table*}

\begin{table*}[!t]
    \caption{Summary of Logistic Sigmoid and Tanh based activation functions.}
    \centering
    {\footnotesize
    \begin{tabular}{p{0.4\textwidth}p{0.07\textwidth}p{0.07\textwidth}p{0.055\textwidth}p{0.156\textwidth}}
        \hline
         Name of AF & Parametric & Monotonic & Smooth & Bounded \\ \hline
         Logistic Sigmoid & No & Yes & Yes & Yes \\
         Tanh & No & Yes & Yes & Yes \\
         Scaled Tanh (sTanh), 1998 \cite{stanh} 
         & Yes & Yes & Yes & Yes \\ 
         Rectified Hyperbolic Secant (ReSech), 2016 \cite{ReSech} & No & No & Yes & Yes \\
         Scaled Sigmoid (sSigmoid), 2016 \cite{rsaf}
         & No & Yes & Yes & Yes \\
         Penalized Tanh (pTanh), 2016 \cite{rsaf}
         & No & Yes & No & Yes \\
         Hexpo, 2017 \cite{hexpo}
         & No & Yes & Yes & Yes \\
         Improved Sigmoid (ISigmoid), 2018 \cite{isigmoid}
         & No & Yes & Yes & No \\
         Sigmoid-Weighted Linear Units (SiLU), 2018 \cite{silu} & No & No & Yes & For negative inputs \\
         Linearly Scaled Hyperbolic Tangent (LiSHT), 2019 \cite{lisht} & No & No & Yes & No \\
         Elliott, 2019 \cite{farzad2019comparative}
         & No & Yes & Yes & Yes \\
         Soft-Root-Sign (SRS), 2020 \cite{srs}
         & Yes & No & Yes & Yes \\
        \hline
    \end{tabular}
    }
    \label{tab:sigmoid}
\end{table*}

% \subsection{Exponential Unit Based Activation Functions}
\textit{Exponential Unit Based Activation Functions:}
The major problem faced by the Logistic Sigmoid and Tanh based AFs is with its saturated output for large positive and negative input. Similarly, the major problem with ReLU based AFs is with the under-utilization of negative values leading to vanishing gradient. In order to cope up with these limitations the exponential function based AFs have been used in the literature. The Exponential Linear Unit (ELU) \cite{elu} based AF utilizes the negative values with the help of the exponential function. Several AFs have been introduced in the literature as the ELU variants which are presented in Section \ref{elu} in detail.

% \subsection{Learning/Adaptive Activation Functions}
\textit{Learning/Adaptive Activation Functions:}
Most of the Sigmoid, Tanh, ReLU, and ELU based AFs are designed manually which might not be able to exploit the data complexity. The learning based adaptive AFs are the recent trends. This class of AFs contains learnable parameters, e.g. Adaptive Piecewise Linear (APL) \cite{apl} and Swish \cite{swish} AFs contain two and one learnable parameters, respectively. 
Recently, several learning based AFs have been proposed as illustrated in Section \ref{learning}.

% \subsection{Miscellaneous Activation Functions}
\textit{Miscellaneous Activation Functions:}
In recent years, many other AFs have also been investigated as presented in Section \ref{other}. These activations include Softplus units, probabilistic functions, polynomial functions, and kernel functions.

Table \ref{tab:evolution_comparison} highlights the advantage and disadvantage of the primary AFs in terms of the diminishing gradients, limited non-linearity, optimization difficulty, computational inefficiency and lack of adaptibility. 
It can be noticed that the Tanh function is computationally inefficient because it involves the computation of exponential multiple times \cite{lecun2015deep}. However, in implementation it can be computed using single exponential with the help of Sigmoid function.
These limitations in the existing AFs have been the driving factors for the development of recent AFs as surveyed in the further sections of this paper.

\section{Logistic Sigmoid and Tanh Based AFs}
\label{sigmoid}
The traditional AFs such as Logistic Sigmoid and Tanh were used very extensively in the early days of neural networks. However, these AFs had shown the hurdle to train the deep networks due to their saturated output. Several attempts have also been made to improve these AFs for different networks. Table \ref{tab:sigmoid} presents the comparison of Logistic Sigmoid and Tanh based AFs in terms of their properties including parametric, monotonic, smooth and bounded.

In order to tackle the limited output range and zero gradient problems of Tanh, a scaled Hyperbolic Tangent (sTanh) is used in \cite{stanh} which is defined as,
\begin{equation}
    sTanh(x) = A\times Tanh(B\times x)
\end{equation}
with the output range in $[-A,A]$.
A Parametric Sigmoid Function (PSF) is proposed as a continuous, differentiable, and bounded function as, 
\begin{equation}
PSF(x) = \frac{1}{(1+e^{-x})^m}    
\end{equation}
where $m$ is a hyperparameter \cite{psf}. The gradient flow is improved for the higher value of $m$. 
The sum of shifted log-sigmoid is also explored as an AF \cite{bdaf} which retains the symmetry in the generated features. 
The Rectified Hyperbolic Secant (ReSech) AF is differentiable, symmetric, and bounded \cite{ReSech} which is given as,
\begin{equation}
ReSech(x) = x\times Sech(x) 
\end{equation}
with the output range in $[-1,1]$.
However, it exhibits the vanishing gradient problem due to saturating behavior for both large positive and large negative inputs.
The training of deep networks become difficult due to the uniform slope of the Logistic Sigmoid and Tanh AFs near the origin \cite{rsaf}. To minimize this limitation, the Scaled Sigmoid (sSigmoid) is defined as,
\begin{equation}
sSigmoid(x) = (4 \times Sigmoid(x) -2)   
\end{equation}
with the output range in $[-2,2]$ and the Penalized Tanh (pTanh) is defined as,
\begin{equation}
    pTanh(x) =  
         \begin{cases}
            Tanh(x), & x \geq 0 \\
            a\times Tanh(x), & x < 0
         \end{cases}
\end{equation}
with the output range in $[-a,1]$ where $a \in (0,1)$. However, sSigmoid and pTanh AFs also suffer from the vanishing gradient problem.
It is noticed that the pTanh AF performs better for Natural Language Processing (NLP) tasks \cite{eger2019time}.

\begin{table*}[!t]
    \caption{Summary of Rectified Linear Unit based activation functions.}
    \centering
    {\footnotesize
    \begin{tabular}{p{0.35\textwidth}p{0.1\textwidth}p{0.14\textwidth}p{0.14\textwidth}p{0.14\textwidth}}
        \hline
         Name & Parametric & Monotonic & Smooth & Bounded \\ \hline
         Rectified Linear Unit (ReLU), 2010 \cite{nair2010rectified} 
         & No \hspace{2cm} & Yes & No & For negative inputs \\
         
         Leaky ReLU (LReLU), 2013 \cite{lrelu} 
         & No & Yes & No & No \\ 
         
         Parametric ReLU (PReLU), 2015 \cite{prelu} 
         & Yes & Yes & No & No \\
         
         Randomized ReLU (RReLU), 2015 \cite{prelu} 
         & No & Yes & No & No \\
         
         Concatenated ReLU (CReLU), 2016 \cite{crelu} 
         & No & Yes & No & For negative inputs \\
         
         Bounded ReLU (BReLU), 2016 \cite{BReLU}
         & No & Yes & No & Yes \\
         
         Parametric Tanh Linear Unit (PTELU), 2017 \cite{ptelu} 
         & Yes & Yes & Yes & For negative inputs \\ 
         
         Flexible ReLU (FReLU), 2018 \cite{frelu}
         & Yes & Yes & No & For negative inputs \\ 
         
         Elastic ReLU (EReLU), 2018 \cite{erelu}
         & No & Yes & No & For negative inputs \\
         
         Randomly Translational ReLU (RTReLU), 2018 \cite{rtrelu}
         & No & Yes & No & For negative inputs \\
         
         Dual ReLU (DualReLU), 2018 \cite{dualrelu}
         & No & Yes & No & No \\
         
         Paired ReLU (PairedReLU), 2018 \cite{pairedrelu}
         & Yes & Yes & No & No \\          
         Average Biased ReLU (ABReLU), 2018 \cite{abrelu}
         & No & Yes & No & For negative inputs \\
         
         Natural-Logarithm (NLReLU), 2019 \cite{nlrelu}
         & No & Yes & No & For negative inputs \\
         
         Multi-bin Trainable Linear Units (MTLU), 2019 \cite{mtlu}
         & Yes & No & No & No \\
         
         Lipschitz ReLU (L-ReLU), 2020 \cite{lstarrelu}
         & Yes & Depends upon $\phi$ and $\eta$ & Depends upon $\phi$ and $\eta$ & Depends upon $\phi$ and $\eta$ \\
         
         \hline
    \end{tabular}
    }
    \label{tab:relu}
\end{table*}

A noisy AF is defined to overcome the vanishing gradient problem \cite{noisy_act}. Due to the added noise the gradients may flow easily even in the saturating regime. 
% The amount of noise is decided based on the saturation level of the non-linearity. 
The vanishing gradient problem is minimized by the Hexpo function \cite{hexpo} which is similar to Tanh with a scaled gradient. It is given as,
\begin{equation}
    Hexpo(x) =  
         \begin{cases}
            -a\times(e^{-x/b}-1), & x \geq 0 \\
            c\times(e^{x/d}-1), & x < 0
         \end{cases}
\end{equation}
in the output range of $[-c,a]$.
The output of the sigmoid function is multiplied with its input in sigmoid-weighted linear unit (SiLU) AF \cite{silu} as
\begin{equation}
SiLU(x) = x\times Sigmoid(x)
\end{equation}
in the output range of $(-0.5,\infty)$. At the same time an improved logistic Sigmoid (ISigmoid) AF \cite{isigmoid} is proposed to solve the vanishing gradient problem of Sigmoid with the help of a piecewise combination of sigmoidal and linear functions. It is defined as,
\begin{equation}
    ISigmoid(x) =  
         \begin{cases}
            \alpha\times(x-a)+Sigmoid(a), & x \geq a \\
            Sigmoid(x), & -a < x < a \\
            \alpha\times(x+a)+Sigmoid(a), & x \leq -a
         \end{cases}
\end{equation}
in the output range of $(-\infty,\infty)$.
The Linearly scaled hyperbolic tangent (LiSHT) AF scales the Tanh in a linear fashion to overcome the vanishing gradient issue \cite{lisht}. The LiSHT can be defined as, 
\begin{equation}
LiSHT(x) = x\times Tanh(x)
\end{equation}
in the output range of $[0,\infty)$.
The LiSHT function is symmetric, but is has the shortcoming of including unbounded and non-negative outputs only. 
The Elliott AF \cite{farzad2019comparative} is similar to Sigmoid function in terms of the characteristics diagram and defined as,
\begin{equation}
Elliott(x) = \frac{0.5\times x}{1+|x|}+0.5
\end{equation}
in the output range of $[0,1]$.
The Soft-Root-Sign (SRS) AF \cite{srs} is defined as,
\begin{equation}
SRS(x) = \frac{x}{\frac{x}{\alpha}+e^{-x/\beta}}
\end{equation}
in the output range of $[\frac{\alpha\times\beta}{\beta-\alpha\times e},\alpha]$ where $\alpha$ and $\beta$ are the learnable parameters. The use of additional parameters increases the complexity of the SRS function.  
Most of the variants of Sigmoid/Tanh AFs have tried to overcome the vanishing gradient issue. However, this issue is still present in most of these AFs.

\section{Rectified Activation Functions}
\label{relu}
A summary of rectified AFs is illustrated in Table \ref{tab:relu}.
Rectified Linear Unit (ReLU) is a simple function which is the identity function for positive input and zero for negative input and given as,
\begin{equation}
    ReLU(x) = max(0,x) =
    \begin{cases}
        x, & \text{if } x\geq 0\\
        0, & \text{otherwise}
    \end{cases}.
\end{equation}
Hence, the range of ReLU is $[0,\infty)$.
The gradient for positive and negative inputs is one and zero, respectively. 
The ReLU function solves the problem of computational complexity of the Logistic Sigmoid and Tanh functions. 
The downside of ReLU is with the vanishing gradient problem for the negative inputs. 
In spite of having the vanishing gradient problem, the ReLU AF has been used very extensively with the deep learning models. 
The advancements in ReLU based AFs are discussed in the rest of this section.

\subsection{On the Non-utilization of Negative Values of ReLU}
Vanishing gradient is the main problem with ReLU AF which is caused due to the non-utilization of negative values. A Leaky Rectified Linear Unit (LReLU) is the extension of ReLU by utilizing the negative values \cite{lrelu}. The LReLU is defined as,
\begin{equation}
    LReLU(x) = \begin{cases}
            x, & x\geq 0\\
            0.01\times x, & x < 0
         \end{cases}
\end{equation}
in the output range of $(-\infty,\infty)$. 
The LReLU has been used in many applications with promising performance.
One major problem associated with LReLU is the finding of the right slope in linear function for negative inputs.
Different slopes might be suited for different problems and different networks. Thus, it is extended to Parametric ReLU (PReLU) by considering the slope for negative input as a trainable parameter \cite{prelu}. The PReLU is given as,
\begin{equation}
    PReLU(x) = \begin{cases}
            x, & x\geq 0\\
            p\times x, & x < 0
         \end{cases}
\end{equation}
in the output range of $(-\infty,\infty)$ where $p$ is the trainable parameter.
However, it can lead to overfitting easily which is the downside of PReLU.
The Maxout layer, which computes the maximum of several linear units, is also used as AF \cite{maxout}. 
Both ReLU and Leaky ReLU can be seen as the special cases of Maxout. 
The randomized ReLU (RReLU) considers the slope of LReLU randomly during training sampled from an uniform distribution $U(l, u)$ \cite{rrelu}. The RReLU is defined as, 
\begin{equation}
    RReLU(x) = \begin{cases}
            x, & x\geq 0\\
            R\times x, & x < 0
         \end{cases}
\end{equation}
in the output range of $(-\infty,\infty)$ where $R \sim{~} U(l, u)$, $l < u$ and $l,u \in [0, 1)$. 
It uses a deterministic value $x/\left (\frac{l+u}{2}\right )$ during test time.

The ReLU is not able to utilize the potential useful information from the negative values. In most of the networks, the feature map given as the input to AF is dense near zero. Thus, a small jitter in the rectification point can lead to difficulty in training. Concatenated ReLU (CReLU) \cite{crelu} concatenates the ReLU's output over original input and negated input. The CReLU can be given as,
\begin{equation}
    CReLU(x) = [ReLU(x), ReLU(-x)]
\end{equation}
in the output range of $[0,\infty)$.
The CReLU is derived from the fact that the lower layer kernels in CNN models form pairs with opposite phases. 
The shifting of the feature map with multiple biases is also performed before the ReLU layer \cite{mba}. 
However, it increases the model complexity as more ReLUs are required. 
A Parametric Tan Hyperbolic Linear Unit (P-TELU) is also used as an AF \cite{ptelu}. The P-TELU is defined as,
\begin{equation}
    PTELU(x) = 
         \begin{cases}
            x, & x\geq 0\\
            \alpha\times\text{Tanh}(\beta \times x), & x < 0
         \end{cases}
\end{equation}
in the output range of $[-\alpha,\infty)$ where $\{\alpha, \beta\} \ge 0$ are the learnable parameters. 

\begin{table*}[!t]
    \caption{Summary of Exponential Linear Unit based activation functions.}
    \small
    \centering
    {\footnotesize
    \begin{tabular}{p{0.4\textwidth}p{0.07\textwidth}p{0.07\textwidth}p{0.05\textwidth}p{0.156\textwidth}}
         \hline
         Name & Parametric & Monotonic & Smooth & Bounded \\ \hline
         Exponential Linear Unit (ELU), 2016 \cite{elu} 
         & Yes & Yes & Yes & For negative inputs \\
         
         Scaled ELU (SELU), 2017 \cite{selu} 
         & Yes & Yes & Yes & For negative inputs \\
         
        Continuously Differentiable ELU (CELU), 2017 \cite{celu} 
         & Yes & Yes & No & For negative inputs \\
         
         Parametric ELU (PELU), 2017 \cite{pelu} 
         & Yes & Yes & No & For negative inputs \\
         
         Multiple PELU (MPELU), 2018 \cite{mpelu} 
         & Yes & Yes & No & For negative inputs \\
         
         Fast ELU (FELU), 2019 \cite{felu} 
         & Yes & Yes & No & For negative inputs \\
         
         Parametric Rectified Exponential Unit (PREU), 2019 \cite{preu} 
         & Yes & No & Yes & For negative inputs \\
         
         Elastic ELU (EELU), 2020 \cite{eelu}
         & Yes & Yes & No & For negative inputs \\
         
         Parametric Deformable ELU (PDELU), 2020 \cite{pdelu} 
         & Yes & Yes & Yes & For negative inputs \\
         \hline
    \end{tabular}
    }
    \label{tab:elu}
\end{table*}

The Flexible ReLU (FReLU) \cite{frelu} captures the negative values with a rectified point which is considered as trainable in the Shifted ReLU \cite{frelu}. The FReLU is given as,
\begin{equation}
FReLU(x) = ReLU(x) + b
\end{equation}
in the output range of $[b,\infty)$.
A similar arrangement is also followed by Random Translation ReLU (RTReLU) \cite{rtrelu} by utilizing an offset, sampled from a Gaussian distribution, given as,
\begin{equation}
    RTReLU(x) = 
         \begin{cases}
            x+a, & x+a > 0\\
            0, & x+a \leq 0
         \end{cases}
\end{equation}
in the output range of $[0,\infty)$ where $a$ is a random number.
At test time, the offset is set to zero. A data dependent Average Biased ReLU (AB-ReLU) \cite{abrelu} is also investigated to tackle the negative values by a horizontal shifting based on the average of features. The ABReLU can be written as,
\begin{equation}
    ABReLU(x) = 
         \begin{cases}
            x-\beta, & x-\beta\geq 0\\
            0, & x-\beta < 0
         \end{cases}
\end{equation}
having the output range in $[0,\infty)$ where $\beta$ is computed as the average of input activation map to the activation function.
The batch dependent threshold for the ReLU is used by the Dynamic ReLU (D-ReLU) \cite{dynamicrelu}.
The Dual ReLU (DualReLU) \cite{dualrelu} is a two dimensional AF for recurrent neural networks. The DualReLU is given as, 
\begin{equation}
    DualReLU(a,b) = \max(0,a) - \max(0,b)
\end{equation}
in the output range of $(-\infty,\infty)$ where $a$ and $b$ are the inputs in different dimensions.
Similar to the CReLU, the PairedReLU AF is used for image super-resolution \cite{pairedrelu}. The PairedReLU is given as,
\begin{equation}
    PairedReLU(x) = [\max(s \times x - \theta, 0), max(s_p \times x - \theta_p, 0)]
\end{equation}
in the output range of $(-\infty,\infty)$. However, the computational complexity of PairedReLU is increased as compared to CReLU.
In another attempt, V-shaped ReLU (vReLU) AF \cite{vrelu} is defined as, 
\begin{equation}
    vReLU(x) = 
         \begin{cases}
            x, & x \geq 0\\
            -x, &  x < 0
         \end{cases}
\end{equation}
having the output range in $[0,\infty]$. The vReLU activation function suffers from the non-symmetric output.
The SignReLU AF utilizes the negative values using the Softsign function \cite{signrelu}. The positive part of SignReLU is the same as the ReLU.

A Displaced ReLU (DisReLU) \cite{displacedrelu} is designed as a generalization of Shifted ReLU \cite{frelu}. The DisReLU displaces the rectification point to consider the negative values, given as,
\begin{equation}
    DisReLU(x) = 
         \begin{cases}
            x, & x \ge -\delta\\
            -\delta, &  x < -\delta
         \end{cases}
\end{equation}
having the output range in $[-\delta,\infty]$. A Bendable Linear Unit (BLU) AF is investigated as,
\begin{equation}
BLU(x) = \beta \times (\sqrt{x^2+1}-1) + x
\end{equation}
where $-1 \le \beta \le 1$ is a learnable parameter to adapt the shape between the identity function and a rectifier function \cite{blu}. 
% The BLU is unbounded for most of the values of $\beta$. 
A Lipschitz ReLU (L-ReLU) AF uses the piecewise linear functions to model the degree of presence and the degree of absence of features \cite{lstarrelu}. The L-ReLU is defined as,
\begin{equation}
    L\text{-}ReLU(x) = 
         \begin{cases}
            \max(\phi(x), 0), & x \geq 0\\
            \min(\eta(x), 0), &  x < 0
         \end{cases}
\end{equation}
where $\phi$ and $\eta$ are non-linear functions. Moreover, the range of L-ReLU also depends upon the values of $\phi$ and $\eta$ functions.

\subsection{On the Limited Non-linearity of ReLU}
S-shaped ReLU (SReLU) increases the non-linearity in ReLU by combining three linear functions with four learnable parameters \cite{SReLU}.
On a similar line, Multi-bin Trainable Linear Unit (MTLU) \cite{mtlu} considers multiple bins to increase the non-linear capacity. The MTLU can be written as,
\begin{equation}
    MTLU(x) = 
         \begin{cases}
            a_0\times x + b_0, & x \leq c_0\\
            a_k\times x + b_k, & c_{k-1} < x \leq c_k\\
            ... & \\
            a_K\times x + b_K, & c_{K-1} < x
         \end{cases}
\end{equation}
having the output range in $(-\infty,\infty)$.
The number of bins and the range of bins are the hyperparameters, whereas the linear function of a bin is trainable (i.e., $a_0,...,a_K$ $b_0,...,b_K$ are the learnable parameters). 
The non-differentiable nature at multiple points is the drawback of the MTLU. 
An Elastic ReLU (EReLU) considers a slope randomly drawn from a uniform distribution during the training for the positive inputs to control the amount of non-linearity \cite{erelu}. The EReLU is defined as,
\begin{equation}
    EReLU(x) = max(R\times x,0)
\end{equation}
in the output range of $[0,\infty)$ where $R$ is a random number. At the test time, the EReLU becomes the identity function for positive inputs.
The Linearized Sigmoidal Activation (LiSHA) function considers three linear functions to increase the non-linearity characteristics \cite{lisha}. It is also extended to adaptive linear sigmoidal AF by learning the slope of upper and lower linear functions. The ReLU is combined with Tanh as Rectified Linear Tanh (ReLTanh) \cite{reltanh} to increase the non-linearity of ReLU and to overcome the vanishing gradient problem of Tanh. However, the ReLTanh is unbounded in both the positive and negative directions. 
Natural-Logarithm ReLU (NLReLU) modifies the ReLU's output for positive inputs using the logarithm function to increase the degree of nonlinearity \cite{nlrelu}. The NLReLU is defined as,
\begin{equation}
    NLReLU(x) = \ln(\beta\times\max(0, x)+1.0)
\end{equation}
having the output range in $[0,\infty)$ where $\beta$ is a constant.
The NLReLU does not affect the negative regime, thus suffers from vanishing gradient.
The concept of Leaky ReLU (LReLU) is further improved to Dynamic ReLU \cite{dynamicreluaccess} by considering a mean square error (MSE) based additional hyperparameter. Thus, it can control the slope of the Dynamic ReLU in every epoch based on the convergence.
A Piecewise Linear Unit (PLU) \cite{plu} is defined as,
\begin{equation}
    PLU(x) = max(\alpha\times(x+c)-c, min(\alpha\times(x-c)+c, x))
\end{equation}
having the output range in $[-\infty,+\infty]$, where $\alpha$ and $c$ are the constants. Basically, the PLU activation function consists of three linear functions in pieces, but continuous. Hence, it avoids the saturation and leads to a good amount of gradient flow through the activation function during backpropagation in order to resolve the vanishing gradient problems of ReLU and Tanh. However, the PLU activation is unbounded in both positive and negative directions.

\subsection{On the Unbounded Output of ReLU}
The unbounded outputs of ReLU and many of its variants may lead to training instability. 
Moreover, the bounded AF is needed for the dedicated hardware based embedded system applications. 
ReLU is extended to Bounded ReLU (BReLU) \cite{BReLU} defined as,
\begin{equation}
    BReLU(x) = \min (\max (0, x) , A)
\end{equation}
having the output range in $[0,A])$.
The training stability is improved in BReLU due to two rectifications (i.e., at $0$ and $A$).
ReLU is a common choice in practice in deep learning. ReLU based AFs are generally efficient. The major drawbacks of ReLU, such as gradient diminishing for negative inputs, limited non-linearity and unboundedness, are improved in the different AFs. However, the ReLU variants are not able to resolve all the issues of ReLU.

\section{Exponential Activation Functions}
\label{elu}
The exponential AFs tackle the gradient diminishing problem of ReLU. 
Table \ref{tab:elu} lists the properties of the exponential AFs. 
The Exponential Linear Unit (ELU) \cite{elu} is given as,
\begin{equation}
    ELU(x) = 
         \begin{cases}
            x, & x > 0\\
            \alpha\times(e^x-1), & x \leq 0
         \end{cases}
\end{equation}
having the output range in $[-1,\infty)$ where $\alpha$ is a learnable parameter.
The ELU function exhibits all the benefits of the ReLU function. The ELU is differentiable, saturates for large negative inputs and reduces the bias shift. The negative saturation regime of ELU adds some robustness to noise as compared to the Leaky ReLU and Parametric ReLU. 
% The ResNet \cite{resnet} has shown a promising performance using the ELU \cite{shah2016deep}. 
% The expensive exponential computation is the major downside of the ELU AF. 
The ELU is extended to Scaled ELU (SELU) \cite{selu} by using a scaling hyperparameter to make the slope larger than one for positive inputs. The SELU can be defined as,
\begin{equation}
    SELU(x) = \lambda\times
         \begin{cases}
            x, & x > 0\\
            \alpha\times(e^x-1), & x \leq 0
         \end{cases}
\end{equation}
having the output range in $[-\lambda,\infty)$ where $\alpha$ is a hyperparameter.
Basically, the SELU induces self-normalization to automatically converge towards zero mean and unit variance. 
% Recently, it has been used in Self-normalizing deep CNNs for speech recognition \cite{huang2019sndcnn}.
The Parametric ELU (PELU) \cite{pelu} changes the saturation point and exponential decay and also regulates the slope of the linear function for the positive inputs for differentiability. The PELU AF can be written as,
\begin{equation}
    PELU(x) = \lambda\times
         \begin{cases}
            \frac{a}{b}\times x, & x \geq 0\\
            a\times(e^{x/b}-1), & x < 0
         \end{cases}
\end{equation}
having $[-a,\infty)$ output range, where $a$ and $b$ are the trainable parameters.
% The CNNs with PELU AF have shown improved performance as compared to ELU for image classification \cite{pelu}. 
The parametric ELU is also explored in Continuously differentiable ELU (CELU) \cite{celu} for the negative inputs. The CELU is given as,
\begin{equation}
    CELU(x) = 
         \begin{cases}
            x, & x \geq 0\\
            \alpha \times (e^{x/\alpha}-1), & x < 0
         \end{cases}
\end{equation}
having the output range in $[-\alpha,\infty)$ where $\alpha$ is a learnable parameter.
The PELU is also extended to multiple PELU (MPELU) \cite{mpelu} by using two learnable parameters to represent MPELU as either rectified, exponential or combined. The MPELU can be expressed as,
\begin{equation}
    MPELU(x) = 
         \begin{cases}
            x, & x > 0\\
            \alpha_c\times(e^{\beta_c\times x}-1), & x \leq 0
         \end{cases}
\end{equation}
having the output range in $[-\alpha_c,\infty)$, where $\alpha_c$ and $\beta_c$ are the trainable parameters.

A soft exponential AF interpolates between the exponential, linear and logarithmic functions using the trainable parameter \cite{softexp}. 
A Shifted ELU (ShELU) AF is also explored as a locally optimal function \cite{shelu}. 
A Parametric Rectified Exponential Unit (PREU) \cite{preu} is designed as, 
\begin{equation}
    PREU(x) = 
         \begin{cases}
            \alpha\times x, & x > 0\\
            \alpha \times x \times e^{\beta \times x}, & x \leq 0
         \end{cases}
\end{equation}
having the output range in $[-1,\infty)$, where $\alpha$ and $\beta$ are the trainable parameters.
The PREU utilizes the negative information near to zero effectively.
The efficiency of ELU is improved in Fast ELU (FELU) AF \cite{felu} with the help of the simple displacement bits and integer algebra operations. The FELU is defined as,
\begin{equation}
    FELU(x) = 
         \begin{cases}
            x, & x > 0\\
            \alpha\times(e^{x/\ln(2)}-1), & x \leq 0
         \end{cases}
\end{equation}
having the output range in $[-\alpha,\infty)$ with $\alpha$ as a learnable parameter.
Recently, the properties of ELU and RELU have been utilized to design an Elastic ELU (EELU) AF \cite{eelu}. The EELU is defined as,
\begin{equation}
    EELU(x) = 
         \begin{cases}
            k\times x, & x > 0\\
            \alpha\times(e^{\beta\times x}-1), & x \leq 0
         \end{cases}
\end{equation}
having the output range in $[-\alpha,\infty)$ where $\alpha$ and $\beta$ are the trainable parameters.
The EELU preserves a small non-zero gradient for the negative input and exhibits an elastic slope for the positive input. 
A Parametric Deformable ELU (PDELU) AF tries to shift the mean value of output closer to zero using the flexible map shape \cite{pdelu}. The PDELU is defined as,
\begin{equation}
    PDELU(x) = 
         \begin{cases}
            x, & x > 0\\
            \alpha\times([1+(1-t)\times x]^{\frac{1}{1-t}}-1), & x \leq 0
         \end{cases}
\end{equation}
having the output range in $[-1,\infty)$ where $\alpha$ is a learnable parameter.
A ReLU-Memristor-like AF (RMAF) \cite{rmaf} uses two hyperparameters to have ReLU like shape for positive input and to give more importance to the negative values near to zero. 
An Exponential Linear Sigmoid SquasHing (ELiSH) is defined in \cite{elish} as,
\begin{equation}
    ELiSH(x) = 
         \begin{cases}
            x/(1 + e^{-x}), & x \ge 0\\
            (e^x - 1)/(1 + e^{-x}), &  x < 0
         \end{cases}
\end{equation}
Moreover, it is also extended to HardELiSH which is a multiplication of HardSigmoid and Linear in the positive part and HardSigmoid and ELU in the negative part. Here, HardSigmoid is defined as,
\begin{equation}
    HardELish(x) = max(0, min(1,(x + 1)/2)).
\end{equation}

The ELU based AFs exploit the negative inputs without compromising with the non-linearity. Some ELU variants also modify the function for positive inputs to make it bounded. 
% Still, the increased computational complexity is the key limiting factor for this class of AFs.

\begin{table*}[!t]
    \caption{Summary of adaptive and learning based activation functions.}
    \centering
    {\footnotesize
    \begin{tabular}{p{0.55\textwidth}p{0.07\textwidth}p{0.07\textwidth}p{0.055\textwidth}p{0.07\textwidth}}
         \hline
         Name & Parametric & Monotonic & Smooth & Bounded \\ \hline
         Adaptive Piecewise Linear Unit (APL), 2015 \cite{apl}
         & Yes & No & No & No \\
         
         Spline AF (SAF), 2016 \cite{saf}
         & Yes & Yes & Yes & No \\
         
         Bi-Modal Derivative Adaptive Activation (BDAA), 2017 \cite{bdaa}
         & Yes & Yes & Yes & Yes \\
         
         Adaptive AF (AAF), 2018 \cite{aaf2017}
         & Yes & Yes & No & No \\
         
         Swish, 2018 \cite{swish}
         & Yes & No & Yes & No \\
         
         ESwish, 2018 \cite{eswish}
         & Yes & No & Yes & No \\
         
         Trainable AF (TAF), 2018 \cite{taf}
         & Yes & No & Yes & No \\
         
         Self-Learnable AF (SLAF), 2019 \cite{slaf}
         & Yes & No & Yes & No \\
         
         Mexican ReLU (MeLU), 2019 \cite{melu}
         & Yes & No & No & No \\
         
         \hline
    \end{tabular}
    }
    \label{tab:adaptive}
\end{table*}

\section{Learning/Adaptive Activation Functions}
\label{learning}
Most of the aforementioned AFs are not adaptive and might not be able to adjust based on the dataset complexity. This problem is tackled using learning/adaptive AFs as summarized in Table \ref{tab:adaptive}. Some of the earlier mentioned AFs are also adaptive, such as PReLU \cite{preu}, SReLU \cite{SReLU}, PTELU \cite{ptelu}, MTLU \cite{mtlu}, PELU \cite{pelu}, MPELU \cite{mpelu}, PREU \cite{preu}, EELU \cite{eelu}, PDELU \cite{pdelu}, SRS \cite{srs}, etc.
% A summary of selected AFs is provided in Table \ref{tab:adaptive}.

The Adaptive Piecewise Linear (APL) is defined as a sum of hinge-shape functions \cite{apl}. It is given as, 
\begin{equation}
APL(x) = \text{max}(0, x) + \sum^S_{s=1}a_s \times  \text{max}(0, b_s-x),    
\end{equation}
where $a$ and $b$ are the trainable parameters and $S$ is a hyperparameter representing the number of hinges. The output range of APL is $[0,\infty)$. Due to the trainable parameters, different neurons can learn different AFs.
% The value of $S$ as $5$ has shown the best performance over the CIFAR10 dataset for image classification.

Ramachandran et al. \cite{swish} have performed an automatic search, which resulted in a Swish AF. It is defined as,
\begin{equation}
    Swish(x) = x \times Sigmoid(\beta \times x)
\end{equation}
where $\beta$ is a learnable parameter. The output range of Swish is $(-\infty,\infty)$. Based on the learnt value of $\beta$ the shape of the Swish AF is adjusted between the linear and ReLU functions. The smaller and higher values of $\beta$ lead towards the linear and ReLU functions, respectively.
Thus, it can control the amount of non-linearity based on the dataset and network complexity.
Swish is also extended to E-Swish by multiplying the Swish with a learnable parameter to control the slope in the positive direction \cite{eswish}. The E-Swish is defined as, 
\begin{equation}
    ESwish(x) = \beta \times x\times Sigmoid(x)
\end{equation}
having the output the range in $(-\infty,\infty)$ and $\beta$ is trainable parameter.
A flatten-T Swish considers zero function for negative inputs similar to the ReLU \cite{flatten_t_swish}. The Adaptive Richard’s Curve weighted Activation (ARiA) is also motivated from Swish and replaces the sigmoidal function with Richard’s Curve \cite{aria}. The ARiA AF uses five hyper-parameters to control the shape of the non-linearity.

% \subsection{Selection/Combination of AFs}
The basic AFs are combined with learnable weights in adaptive AFs \cite{aaf2017}. 
The Adaptive AF (AAF) designed over PReLU \cite{prelu} and PELU \cite{pelu} is given as,
\begin{equation}
    AAF(x) = \sigma(w\times x)\times PRELU(x) + (1 - \sigma(w\times x))\times PELU(x)
\end{equation}
having the output range in $[0,1]$,
where $\sigma$ is the sigmoidal function and $w$ is a learnable parameter. 
In practice, AAF is costly as multiple AFs are involved.
In \cite{aaf2016}, the AF for each neuron is selected from a library of AFs.
In \cite{manessi2018learning}, different combinations of the identity function, ReLU, and Tanh are learnt automatically.
In another attempt, an Adaptive Blending Unit (ABU) is defined to allow the networks to learn its preferred AFs \cite{abu}. The ABU combines a set of AFs with trainable weights. A Lookup Table Unit (LuTU) function \cite{lutu} uses a single period cosine mask based smoothing and linear interpolation using a set of anchor points. 
% It is a complex function with multiple local minima.
Activation ensembles are used at each layer in \cite{activationensembles} with the contribution of each AF controlled by the trainable weights.
Similarly, the Self-Learnable AF (SLAF) computes the sum of the different functions in an ensemble with the learnt coefficients \cite{slaf}. The SLAF can be expressed as,
\begin{equation}
    SLAF(x) = \sum_{i=0}^{N-1} a_i \times x^i
\end{equation}
in the output range of $(-\infty, \infty)$, where $a_i$ is the trainable parameter.
% A CNN ensemble is defined with each CNN trained using the distinct AFs \cite{melu}. 
A Mexican ReLU (MeLU) AF is proposed in \cite{melu} by using a “Mexican hat type” function and given as,
\begin{equation}
MeLU(x) = PReLU(x)+\sum_{j=1}^{k}{c_j\times\max(\lambda_j-|x-a_j|,0)}    
\end{equation}
in the output range of $(-\infty, \infty)$, where $c_j$ is the trainable parameter and $\lambda_j$ \& $a_j$ are the real numbers.

% \subsection{Other Adaptive/Learnt AFs}
A cubic spline interpolation is also used to learn the AF from data \cite{saf} which is given as,
\begin{equation}
    SAF(x) = \Phi(s;\textbf{q})
\end{equation}
having the output range in $(-\infty,\infty)$ where $\Phi(.)$ is parameterized by a vector $\textbf{q}$ cubic in nature.
Fourier series basis expansion is used for nonparametrically learning AFs (NPF) \cite{npf}. 
Hyperactivations utilize a hypernetwork on top of an activation network, which are used to explore the AFs search space \cite{hyperactivations}. A shallow neural network is used in the activation network to produce the output for each input, whereas a neural network is used in the hypernetwork to produce weights for another network. 
A bi-modal derivative adaptive activation (BDAA) function uses twin maxima derivative sigmoidal function \cite{bdaa} by controlling the maxima's position with an adaptive parameter. The BDAA is given as,
\begin{equation}
    BDAA(x) = \frac{1}{2}\times\left(\frac{1}{1+e^{-x}}-\frac{1}{1+e^{-x-a}}\right)
\end{equation}
in the output range of $[0,1]$ where $a$ is a learnable parameter.
The authors have exploited the Bi-modal derivatives on four AFs.
Linear regression is used in \cite{taf} to train AF for each neuron which results in different AFs for the different neurons.
The TAF is defined as,
\begin{equation}
    TAF(x) = \sqrt{(x-a)^2+b^2}
\end{equation}
in the output range of $[b,\infty)$, where $a$ and $b$ are the trainable parameters.
% by considering the sum of the inputs and target outputs over the original training set. This process creates . 
Recently, a trainable parameter was used in different non-adaptive AFs such as Sigmoid, Tanh, and ReLU to make it adaptive \cite{aaf2020}.
 
\textcolor{red}{The adaptive and trainable AFs are the recent trend to adjust the non-linearity based on the data and network complexity. However, the minimal burden is increased in terms of the increased number of parameters. Though the complexity of tunable AFs is relatively increased w.r.t. non-tunable AFs, it is negligible w.r.t. all parameters of the entire network in practice. The same is also observed experimentally as reported in Table \ref{tab:results_cifar100_time} in terms of the training time.}

\section{Miscellaneous Activation Functions}
\label{other}
This section covers other attempts in AFs such as Softplus, Probabilistic, Polynomial, Subnetwork and Kernel.

\subsection{Softplus Activation Functions}
The softplus function \cite{softplus_original} was proposed in 2001 as $\log(e^x+1)$ and mostly used in statistical applications. After the breakthrough of deep learning the softmax function is used as the AF \cite{glorot2011deep}. 
Softmax function produces the categorical probability distribution equivalent output. 
Softplus unit based AF is also used in deep neural networks \cite{softplus}. The smooth nature of the Softplus facilitates the differentiability. The noisy softplus AF \cite{noisy_softplus} is suitable for the spiking neural networks (SNNs). 
% The non zero-centric output of noisy softplus is the major drawback.
A Softplus Linear Unit (SLU) is also proposed by considering softplus with rectified unit \cite{slu}. The SLU AF is defined as,
\begin{equation}
    SLU(x) = 
         \begin{cases}
            \alpha\times x, & x \geq 0\\
            \beta\times\log(e^x + 1) - \gamma, &  x < 0
         \end{cases}
\end{equation}
where $\alpha$, $\beta$ and $\gamma$ are the trainable parameters with $\alpha$ controlling the slope in the positive direction, $\beta$ controlling the saturation points in the negative direction and $\gamma$ controlling the offset in the negative direction w.r.t. the horizontal axis.
The Rectified Softplus (ReSP) AF introduces the rectification for positive input in Softplus activation \cite{rectifiedsoftplus}. 
In order to make the softplus function to follow the zero mean, a shifting and scaling of the outputs is performed in \cite{isa}. A Rand Softplus (RSP) AF models the stochasticity-adaptability of biological neurons as,
\begin{equation}
RSP(x) = (1 - \rho)\times \max(0, x) + \rho\times \log(1 + e^x)    
\end{equation}
where $\rho$ is a stochastic hyperparameter \cite{rsp}. It improves the capability of the network towards the noise. The softplus function is also used with Tanh function in Mish activation function \cite{mish}, which is given as,
\begin{equation}
    Mish(x) = 
    x\times Tanh(Softplus(x)).
\end{equation}
The Mish is a non-monotonic and smooth AF. It has recently been used by the YOLOv4 model for object detection \cite{bochkovskiy2020yolov4}. However, the increased complexity in Mish due to the multiple functions can be a limitation for the deep networks.

\subsection{Probabilistic Activation Functions}
So far, stochastic AFs have not been much explored due to expensive sampling processes. Few AFs exist in this category such as Randomized ReLU (RReLU) \cite{rrelu}, Elastic ReLU (EReLU) \cite{erelu}, Randomly Translational ReLU (RTReLU) \cite{rtrelu} and Gaussian Error Linear Unit (GELU) \cite{gelu}. 
GELU \cite{gelu} considers nonlinearity as the stochastic regularization driven transformation and defined as,
\begin{equation}
GELU(x) = x\times P(X \le x).    
\end{equation}
where $P$ is the probability.
The complexity of GELU increases due to use of probabilistic nature. The GELU is also extended to the Symmetrical Gaussian Error Linear Unit (SGELU) \cite{sgelu} to enhance its ability of bidirectional convergence. 
% The shape of SGELU is similar to LiSHT \cite{lisht}.
Doubly truncated Gaussian distributions \cite{trug} is a family of nonlinearities which can generate different AFs such as Sigmoid, Tanh and ReLU by setting the appropriate truncation points.
Probabilistic AF (ProbAct) introduces the adaptable and trainable variance in the ReLU's output \cite{probact}. It leads to the generalization of the models. However, all other drawbacks of ReLU exist with ProbAct also. 

\subsection{Polynomial Activation Functions}
Smooth Adaptive AF (SAAF) is defined as the piecewise polynomial function \cite{saaf}. 
Two power functions symmetric to the linear part of ReLU are combined in \cite{berradi2018symmetric} to improve the performance of ReLU.
A piecewise polynomial approximation based AF is also learnt from the data \cite{lopez2019piecewise}. This activation leads to the light-weight models suitable for the FPGAs and microcontrollers. The AF is also treated as the cumulative distribution function \cite{farhadi2019activation}.
The ReLU is also extended to a Rectified Power Unit (RePU) for positive inputs as,
\begin{equation}
    RePU(x) = 
         \begin{cases}
            x^s, & x \geq 0\\
            0, &  x < 0
         \end{cases}
\end{equation}
where $s$ is a hyperparameter \cite{repu}. The RePU is suitable for smoother gradients near zero. However, vanishing gradient, unbounded and asymmetric nature are the downsides of RePU.
The rational function of polynomials is better suited as compared to the polynomial functions in order to approximate the ReLU \cite{telgarsky2017neural}. 
Recently, a Padé approximation is used to develop a non-smooth Padé Activation Unit (PAU) \cite{pau} as,
\begin{equation}
 PAU(x) = \frac{P(x)}{Q(x)}
\end{equation}
where $P(x)$ and $Q(x)$ are two polynomials of order $m$ and $n$, respectively. The PAUs can approximate the commonly used hand-designed AFs. Moreover, it can also learn the new AFs with compact representations. 
% The denominator function of PAU leads to the poles of PAU on the imaginary axis, which can lead to the costly gradient computation due to the non-smooth AF.
Recently, a Rational AF (RAF) \cite{raf} was proposed to tackle the problem of non-smooth nature of the PAU function.

\subsection{Activations as a Subnetwork}
A Variable AF (VAF) is used as a subnetwork of ReLUs \cite{vaf}. It uses the ensemble of ReLUs in a subnetwork using learnable parameters. In a very similar approach, the maximum of multiple linear functions is used in the Dynamic ReLU (DY-ReLU) \cite{dyrelu}. In Wide Hidden Expansion (WHE) \cite{whe}, each WHE intermediate channel is followed by one AF before connecting to the output channel to increase the non-linearity of the network.
An AF Unit (AFU) \cite{afu} uses a small neural network to model the activation. All neurons in the original network share the weights in AFU. The advantage of the AFU is that different AFs can be learnt by different layers.

\subsection{Kernel Activation Functions}
A Kernel-based non-parametric AF (KAF) \cite{kaf} uses an inexpensive kernel expansion to make the activation flexible. The KAF is further extended to multikernel AFs (multi-KAF) \cite{multikaf}. Several AFs are also introduced for complex valued neural networks \cite{cvaf}, \cite{wlkaf}, \cite{kobayashi2017singularities}.

% \subsection{Genetic Algorithm Based Activation Fuctions}
% An Exponential Linear Sigmoid SquasHing (ELiSH) is defined in \cite{elish} as $x/(1 + e^{-x})$ for $x\ge0$ and $(e^x - 1)/(1 + e^{-x})$ for $x<0$ which is suitable for genetic algorithm. Moreover, it is also extended to HardELiSH which is used to improve the performance of the U-Net model for MRI brain tumor segmentation \cite{salih2019enhancement}.
% The genetic algorithm based AF is also used in Neuroevolutionary based convolutional neural network \cite{zahedinasab2020neuroevolutionary}.

% \subsection{Other Activation Functions}
% A non-sigmoidal, bounded, continuous, differentiable, non-constant and non-polynomial AF is used in \cite{chandra2015non}. 
% A Parametric Algebraic Activation (PAA) function is introduced as the family of the S-shaped curves \cite{paa}. 
% A Self-Normalizing Activation (SNA) function is also used to improve the memory capacity, especially for the echo state networks \cite{sna}.
% Some other AFs are piecewise linear AFs \cite{guo2018multistability}, and interval type-2 fuzzy rectifying unit based AF \cite{beke2019interval}.

\begin{table*}[!t]
\caption{Summary of the existing state-of-the-art activation functions.}
\centering
{\footnotesize
\begin{tabular}{p{0.18\textwidth}p{0.17\textwidth}p{0.2\textwidth}p{0.32\textwidth}}
\hline
Activation & Models & Datasets & Insights and Remarks \\\hline
\multicolumn{4}{c}{\textbf{On Image Datasets}}\\\hline
Wide Hidden Expansion (WHE) - 2020 \cite{whe} & ResNet, SENet, and MobileNet & CIFAR100 and ImageNet classification, Pascal VOC 2007 and COCO detection & Upto 2\% higher Top-1 accuracy than baseline models of recognition and detection \\\hline
Soft-Root-Sign (SRS) - 2020 \cite{srs} & VGG and MobileNet & CIFAR10 and CIFAR100 classification & The SRS is better with MobileNet over both datasets and with VGG over CIFAR100. The LReLU is better with VGG over CIFAR10. \\\hline
Relu-Memristor-Like AF (RMAF) - 2020 \cite{rmaf} & ResNet, AlexNet, SqueezeNet, and DenseNet & CIFAR10, CIFAR100, MNIST and ImageNet classification & The RMAF performs better than the ReLU, ELU, SELU, PReLU, Tanh and Swish. \\\hline
Parametric Deformable ELU (PDELU) - 2020 \cite{pdelu} & NIN and ResNet & CIFAR10 and CIFAR100 classification & The PDELU performs better than the ReLU, ELU and FReLU. \\\hline
Pade Activation Unit (PAU) - 2020 \cite{pau} & VGG8, MobileNetV2, ResNet and DenseNet & MNIST, Fashion-MNIST, CIFAR10 and ImageNet classification & The PAU encode AFs as rational functions and performs better than many existing AFs. \\\hline
Elastic Exponential Linear Unit (EELU) - 2020 \cite{eelu} & A simple CNN model and VGG16 & CIFAR10, CIFAR100, ImageNet, and Tiny ImageNet classification & The EELU shows better results than the ReLU, ELU, EPReLU and Swish. \\\hline
Dynamic ReLU (DY-ReLU) - 2020 \cite{dyrelu} & MobileNetV2 & ImageNet classification and COCO detection & The DY-ReLU is suitable for light-weight networks. \\\hline
Variable AF (VAF) - 2019 \cite{vaf} & Shallow CNN models & MNIST, Fashion MNIST and CIFAR10 classification & The VAF shows promising performance. \\\hline
Multi-bin Trainable Linear Unit (MTLU) - 2019 \cite{mtlu} & FDnet and FSRnet & Image denoising and
Super-resolution & The MTLU is significantly faster having comparable results with the state-of-the-arts. \\\hline
Swish - 2018 \cite{swish} & MobileNet, ResNet, WRN and DenseNet & CIFAR10, CIFAR100 and ImageNet classification & The learnable parameter in Swish leads to improved performance than Softplus. \\\hline
\multicolumn{4}{c}{\textbf{On Time Series Datasets}} \\\hline
Variable AF (VAF) - 2019 \cite{vaf} & Multi-Layered Neural Network & Regression tasks (Kinematics, Energy Cooling, Yatch, etc.) & Better performance over Kinematics, Energy Cooling and Yatch datasets. \\\hline
Self-Learnable AFs (SLAF) - 2019 \cite{slaf} & Multi-Layered Neural Network & Boston Housing and Learning Sparse Polynomial regression & The newer parameter space makes the optimization easier. \\\hline
\multicolumn{4}{c}{\textbf{On Text Datasets}} \\\hline
Soft-Root-Sign (SRS) - 2020 \cite{srs} & A 6 layer transformer network & IWSLT 2016 German-English translation & The SRS is better over tst2011 and tst2012 test sets, whereas the SELU and LReLU are better over tst2013 and tst2014 test sets, respectively. \\\hline
Swish - 2018 \cite{swish} & A 12 layer transformer network & WMT 2014 English-German dataset & The performance of Swish is comparable to state-of-the-arts. \\\hline
PenalizedTanh - 2018 \cite{eger2019time} & MLP, CNN and RNN & Sentence classification, Document classification and Sentence tagging & The PenalizedTanh exhibits the stability across the different tasks in contrast to the Swish function. \\\hline
\multicolumn{4}{c}{\textbf{On Signal Datasets}} \\\hline
Rectified Linear Tanh (ReLTanh) - 2019 \cite{reltanh} & Stacked autoencoder (SAE) based DNN & Vibration signals for rotating machinery fault diagnosis & The ReLTanh leads to larger gradients for faster learning and reduces the vanishing gradient. \\\hline
\multicolumn{4}{c}{\textbf{On Game Datasets}} \\\hline
Sigmoid-weighted Linear Unit (SiLU) - 2018 \cite{silu} & Deep reinforcement learning algorithm & SZ-Tetris, $10\times10$ Tetris, and Atari 2600 games & The SiLU AF outperforms the ReLU function for reinforcement learning. \\\hline
\end{tabular}
}
\label{tab:act_summary}
\end{table*}

\section{Aspects of Activation Functions}
\label{aspects}
This section summarizes the effect of weight initialization, understanding of AFs and suitability with different types of data.
The learning of the network speeds up drastically by using the orthogonal weight initialization based on the dynamical isometry \cite{pennington2017resurrecting}. 
% The sigmoidal function is better suited for dynamical isometry.
A set of conditions in parameter initialization also boosts the performance of networks with sigmoidal activations \cite{sansone2017training}.
The symmetric probability distribution based weights and biases initialization leads the network to suffer with the dying ReLU problem. However, the asymmetric initialization resolves the dying ReLU problem \cite{dyingrelu}. The over-parameterization during initialization also benefits in the training \cite{arpit2019benefits}. The data-dependent weight initialization using a subset of data minimizes the issues of the ReLU \cite{aguirre2019improving}, whereas an initial parameter sharing based initialization guarantees the dynamical isometry for the ReLU \cite{burkholz2019initialization}.

Several researchers have tried to understand the working and impact of AFs through different strategies.
The lower and upper bounds are established for network complexity to realize that the ReLU in deep networks approximates the smooth functions more efficiently as compared to shallow networks \cite{yarotsky2017error}. 
A ReLU network with only one hidden layer is trained to reach the global optimum in polynomial time even with exponentially growing input dimension \cite{arora2016understanding}.
The ReLU type AF based neural networks produce the overconfident predictions far away from the training data \cite{hein2019relu}. However, this can be resolved by employing adversarial confidence enhanced training. 
A Gaussian margin driven time and accuracy tradeoff analysis is also done on the ReLU's learning \cite{goel2019time}.
The singular values for ReLU layers are analyzed to understand the interaction of ReLU with the linear components \cite{dittmer2019singular}.
The approximation of Gaussian posterior distribution over the ReLU network weight's fixes the overconfidence problem \cite{kristiadi2020being}. 

Despite most of the AFs are tested over image data, there are few research papers dealing with the AFs over other types of data. Table \ref{tab:act_summary} summarizes the insights and remarks of state-of-the-art AFs for various networks and datasets.
% , including image, series, text, signal and game. 
% Note that the performance metric is not compared in Table \ref{tab:act_summary} because the results in these papers are computed under different experimental settings. 

\section{Performance Comparison and Analysis}
\label{performance}
This survey is compared with the existing survey/performance analysis and the experimental performance analysis of selected AFs is performed over Image, Text and Speech data.

\subsection{Comparison with Existing Survey/Performance Analysis}
A performance analysis of AFs was conducted using multilayer perceptron network in \cite{karlik2011performance}. Among compared AFs, the Tanh has shown better performance.
A comparative performance analysis of different AFs suggests an Elliott function as better suited for classification using LSTM networks \cite{farzad2019comparative}. The ELU outperforms the ReLU, LReLU, and SELU AFs over MNIST classification task using Deep Neural Networks \cite{alcantara2017empirical}.
As per \cite{vydana2017investigative}, the ELU is reported in \cite{alcantara2017empirical} to outperform the ReLU, LReLU, PReLU and PELU over sufficiently large datasets for speech recognition. However, for smaller datasets, the ReLU is preferred. A similar trend is also reported in \cite{pedamonti2018comparison} with a note that the ELU and SELU AFs exhibit faster convergence as compared to the ReLU and LReLU AFs.
In \cite{nwankpa2018activation}, 21 AFs are listed without experimental results comparison. In contrast to \cite{nwankpa2018activation}, this paper presents a comprehensive survey of AFs. The ReLU based deep networks perform superior or mildly worse than the spline methods \cite{eckle2019comparison}. A review of adaptive functions is conducted in \cite{lau2018review} by considering 9 functions, including Sigmoid, Tanh, PReLU, and adaptTanh.
In \cite{dubey2019comparative}, the comparison between ReLU and LReLU is performed using CNN on MNIST dataset. An empirical study is also done for the variations of ReLU activation by generalizing it with the help of parameters \cite{banerjee2019empirical}.
The comparison of AFs is also performed for generalized learning vector quantization \cite{villmann2019activation}. 
The ReLU activation has performed better for object, face, and text datasets \cite{castaneda2019evaluation}. However, the SELU and Maxout have performed better for medical and sound datasets, respectively \cite{castaneda2019evaluation}.
The piecewise AF is better suited for facial expression recognition in \cite{wang2020influence}. 
A survey of adaptive AFs is conducted in \cite{apicella2020survey} without experimental comparison. The evaluation of seven AFs is conducted in \cite{szandalareview} using a simple network over CIFAR10 dataset, whereas in our survey we cover different AFs and also perform the experimental comparison. 

\begin{table*}[!t]
    \caption{Comparison of this survey with the existing surveys and performance evaluations.}
    \centering
    {\footnotesize
    \begin{tabular}{p{0.1\textwidth}p{0.15\textwidth}p{0.2\textwidth}p{0.17\textwidth}p{0.22\textwidth}}
        \hline
        Method & Models & Activations & Datasets & Remarks \\\hline
        Karlik and Olgac \cite{karlik2011performance} & Multilayer Perceptron (MLP) & 5 AFs, including Bi-polar sigmoid, Uni-polar sigmoid, Tanh, etc. & Classification & The Tanh performs better compared to other traditional AFs. \\\hline
        Vydana and Vuppala (2017) \cite{vydana2017investigative} & Hidden Markov Model-Deep Neural Network (HMM-DNN) & 5 AFs, including ReLU, LReLU, PReLU, ELU, and PELU & TIMIT and WSJ speech recognition & The ELU is better over sufficiently larger size datasets. However, the ReLU is preferred for smaller datasets. \\\hline
        Alcantara (2017) \cite{alcantara2017empirical} & A neural network with 2 hidden layers having 100 neurons/layer & 4 AFs, including ReLU, LReLU, ELU, and SELU & MNIST classification & The ELU AF outperforms others. \\\hline
        Pedamonti (2018) \cite{pedamonti2018comparison} & A neural network with 2 hidden layers having 100 neurons/layer & 5 AFs, including Sigmoid, ReLU, LReLU, ELU, and SELU & MNIST classification & The ELU and SELU AFs exhibit the faster convergence as compared to the ReLU and LReLU AFs. \\\hline
        Lau and Lim (2018) \cite{lau2018review} & Deep Neural Network (DNN) & ReLU and Adaptive ReLU & MNIST classification & The adaptive AFs improve the generalization of the network.\\\hline
        Farzad et al. (2019) \cite{farzad2019comparative} & Long Short Term Memory (LSTM) & 23 AFs, including Elliott, Gaussian, Logarithmic, Loglog, etc. & IMDB, Movie Review, MNIST classification & Elliott  function is better  suited to the LSTM  network. \\\hline
        Dubey and Jain (2019) \cite{dubey2019comparative} & Simple Convolutional Neural Network (CNN) & 2 AFs, including ReLU and Leaky ReLU & MNIST classification & The ReLU performed better than Leaky ReLU (LReLU). \\\hline
        Banerjee et al. (2019) \cite{banerjee2019empirical} & Convolutional Neural Network (CNN) & Generalized ReLU & MNIST classification & Network learns the parameters for different ReLU variations. \\\hline
        Villmann et al. (2019) \cite{villmann2019activation} & Generalized learning vector quantization (GLVQ) & 12 AFs, including Sigmoid, Swish, ReLU, Softplus, etc. & Tecator, Indian Pine and Wisconsin-Breast-Cancer classification & The Sigmoid, Swish and Softplus AFs are better suited with GLVQ. \\\hline
        Castaneda et al. (2019) \cite{castaneda2019evaluation} & 6 different models for different applications & 3 AFs, including ReLU, SELU and Maxout & Object, Face, Text, Medical and Sound datasets & The ReLU is better for object, face and text datasets, whereas SELU and Maxout are better for medical and sound datasets, respectively. \\\hline
        
        Wang et al. (2020) \cite{wang2020influence} & Inception-v3 model & 6 AFs, including Sigmoid, Tanh, ReLu, etc. & JAFFE and FER2013 facial expression recognition & The combination of log, softdesign and ReLU AFs provides improved performance.\\\hline
        Szandala (2020) \cite{szandalareview} & A simple network & 7 AFs, including Sigmoid, Tanh, ReLU, LReLU, Swish, etc. & CIFAR10 classification & The LReLU performs better. The ReLU is efficient.\\\hline
        Our survey and performance analysis & MobileNet, VGG, GoogLeNet, ResNet, SENet, DenseNet, etc. & Exhaustive list of AFs, including performance analysis over $18$ state-of-the-art activations & CIFAR10 classification, Language translation, Speech recognition & A classification to categorize and analyze the AFs and a performance comparison of the state-of-the-art activations. \\\hline
    \end{tabular}
    }
    \label{tab:existing_survey}
\end{table*}

A summary of the comparison with existing surveys and performance analysis of AF is shown in Table \ref{tab:existing_survey}. Following are the observations:
\begin{itemize}
    \item 
    % (a) 
    This survey presents a detailed classification to cover the wide range of AFs as compared to the existing surveys and performance analysis.
    \item 
    % (b) 
    This survey covers exhaustive state-of-the-art AFs to date, whereas the existing survey/performance analysis covers either a limited number of AFs or only basic AFs.
    \item 
    % (c) 
    The performance analysis conducted in this paper considers a wide range of neural networks over different types of data for eighteen AFs, whereas the existing analysis is limited to a single type of data and network. 
    \item 
    % (d) 
    This survey highlights the trends to help the researchers to further explore the better AFs and practitioners to choose based on the data and network types. 
\end{itemize}

\begin{table*}[!t]
    \caption{Experimental results comparison over CIFAR10 dataset.}
    \centering
    {\footnotesize
    \begin{tabular}{p{0.128\textwidth}p{0.11\textwidth}p{0.11\textwidth}p{0.11\textwidth}p{0.11\textwidth}p{0.11\textwidth}p{0.11\textwidth}}
    \hline
    Accuracy & \multicolumn{6}{c}{CNN Models} \\\cline{2-7}
    Activations & MobileNet & VGG16 & GoogleNet & ResNet50 & SENet18 & DenseNet121 \\\hline
    Sigmoid	&	88.60	$\pm$	0.17	&	87.69	$\pm$	0.49	&	87.33	$\pm$	2.48	&	80.13	$\pm$	3.33	&	90.29	$\pm$	0.29	&	89.92	$\pm$	1.96	\\
    Tanh	&	87.21	$\pm$	0.24	&	90.49	$\pm$	0.11	&	90.16	$\pm$	1.86	&	89.09	$\pm$	1.47	&	90.44	$\pm$	0.09	&	91.80	$\pm$	0.69	\\
    Elliott	\cite{farzad2019comparative} &	88.48	$\pm$	0.18	&	87.94	$\pm$	0.49	&	89.84	$\pm$	3.43	&	81.60	$\pm$	3.91	&	90.25	$\pm$	0.25	&	91.53	$\pm$	1.04	\\
    ReLU \cite{alexnet}	&	90.10	$\pm$	0.22	&	\textit{92.84}	$\pm$	0.19	&	\textit{93.43}	$\pm$	0.48	&	93.74	$\pm$	0.34	&	93.70	$\pm$	0.16	&	\textbf{93.96}	$\pm$	0.51	\\
    LReLU \cite{nair2010rectified}	&	90.10	$\pm$	0.19	&	91.09	$\pm$	0.09	&	89.28	$\pm$	0.82	&	\textit{93.83}	$\pm$	0.42	&	93.66	$\pm$	0.19	&	93.85	$\pm$	0.48	\\
    PReLU \cite{prelu}	&	90.43	$\pm$	0.18	&	92.19	$\pm$	0.08	&	92.85	$\pm$	0.55	&	92.99	$\pm$	0.62	&	92.76	$\pm$	0.26	&	92.82	$\pm$	0.63	\\
    ELU \cite{elu}	&	90.92	$\pm$	0.25	&	88.55	$\pm$	1.17	&	92.47	$\pm$	0.76	&	93.53	$\pm$	0.66	&	93.39	$\pm$	0.20	&	92.89	$\pm$	0.62	\\
    SELU \cite{selu}	&	90.11	$\pm$	0.32	&	92.25	$\pm$	0.28	&	91.87	$\pm$	0.84	&	93.53	$\pm$	0.52	&	89.96	$\pm$	0.31	&	92.71	$\pm$	0.73	\\
    GELU \cite{gelu}	&	90.71	$\pm$	0.20	&	92.42	$\pm$	0.09	&	93.16	$\pm$	0.61	&	93.81	$\pm$	0.46	&	\textbf{93.72}	$\pm$	0.18	&	\textit{93.90}	$\pm$	0.41	\\
    CELU \cite{celu}	&	\textit{91.04}	$\pm$	0.17	&	88.11	$\pm$	0.14	&	92.60	$\pm$	0.60	&	\textbf{94.09}	$\pm$	0.17	&	91.63	$\pm$	0.22	&	93.46	$\pm$	0.35	\\
    Softplus \cite{softplus}	&	\textbf{91.05}	$\pm$	0.22	&	92.69	$\pm$	0.20	&	92.66	$\pm$	0.66	&	93.34	$\pm$	0.65	&	93.25	$\pm$	0.11	&	93.07	$\pm$	0.70	\\
    Swish \cite{swish}	&	90.66	$\pm$	0.34	&	92.32	$\pm$	0.20	&	92.68	$\pm$	0.53	&	93.02	$\pm$	0.85	&	93.24	$\pm$	0.19	&	93.16	$\pm$	0.51	\\
    ABReLU \cite{abrelu}	&	88.97	$\pm$	0.47	&	92.36	$\pm$	0.15	&	93.34	$\pm$	0.23	&	93.29	$\pm$	0.52	&	93.35	$\pm$	0.14	&	93.26	$\pm$	0.55	\\
    LiSHT \cite{lisht}	&	86.53	$\pm$	0.49	&	89.83	$\pm$	0.28	&	90.27	$\pm$	0.80	&	90.89	$\pm$	0.66	&	90.25	$\pm$	0.84	&	87.91	$\pm$	0.93	\\
    SRS \cite{srs}	&	89.43	$\pm$	0.81	&	92.06	$\pm$	0.26	&	91.36	$\pm$	1.19	&	92.28	$\pm$	0.48	&	78.05	$\pm$	1.37	&	90.64	$\pm$	1.93	\\
    Mish \cite{mish}	&	90.82	$\pm$	0.15	&	\textbf{92.85}	$\pm$	0.25	&	93.29	$\pm$	0.61	&	93.69	$\pm$	0.63	&	93.66	$\pm$	0.12	&	93.62	$\pm$	0.62	\\
    PAU \cite{pau}	&	90.67	$\pm$	0.17	&	92.00	$\pm$	0.26	&	92.80	$\pm$	0.65	&	93.67	$\pm$	0.52	&	93.08	$\pm$	0.20	&	93.05	$\pm$	0.53	\\
    PDELU \cite{pdelu}	&	90.18	$\pm$	0.19	&	92.80	$\pm$	0.13	&	\textbf{93.49}	$\pm$	0.30	&	93.42	$\pm$	0.71	&	\textit{93.71}	$\pm$	0.07	&	\textbf{93.96}	$\pm$	0.59	\\\hline
    \end{tabular}
    }
    \label{tab:results_cifar}
\end{table*}

\begin{table*}[!t]
    \caption{Experimental results comparison over CIFAR100 dataset. }
    \centering
    {\footnotesize
    \begin{tabular}{p{0.128\textwidth}p{0.11\textwidth}p{0.11\textwidth}p{0.11\textwidth}p{0.11\textwidth}p{0.11\textwidth}p{0.11\textwidth}}
    \hline
    Accuracy & \multicolumn{6}{c}{CNN Models} \\\cline{2-7}
    Activations & MobileNet & VGG16 & GoogleNet & ResNet50 & SENet18 & DenseNet121 \\\hline
    Sigmoid	&	61.88	$\pm$	0.18	&	37.75	$\pm$	0.59	&	70.31	$\pm$	0.54	&	46.78	$\pm$	5.42	&	66.17	$\pm$	1.16	&	68.31	$\pm$	2.41	\\
    Tanh	&	53.10	$\pm$	0.51	&	58.43	$\pm$	0.38	&	67.66	$\pm$	2.32	&	64.32	$\pm$	1.69	&	60.13	$\pm$	1.86	&	69.53	$\pm$	1.68	\\
    Elliott \cite{farzad2019comparative}	&	60.70	$\pm$	0.34	&	33.20	$\pm$	0.97	&	64.85	$\pm$	6.28	&	49.88	$\pm$	4.03	&	66.30	$\pm$	0.28	&	69.58	$\pm$	2.40	\\
    ReLU \cite{alexnet}	&	61.33	$\pm$	0.34	&	67.47	$\pm$	0.44	&	74.05	$\pm$	1.69	&	71.96	$\pm$	0.94	&	70.45	$\pm$	0.73	&	72.99	$\pm$	1.35	\\
    LReLU \cite{nair2010rectified}	&	61.13	$\pm$	0.41	&	65.72	$\pm$	0.14	&	63.79	$\pm$	2.38	&	\textbf{72.77}	$\pm$	0.49	&	70.58	$\pm$	0.45	&	73.33	$\pm$	1.25	\\
    PReLU \cite{prelu}	&	59.86	$\pm$	0.35	&	65.26	$\pm$	0.40	&	69.57	$\pm$	1.50	&	71.08	$\pm$	1.70	&	69.77	$\pm$	0.48	&	68.23	$\pm$	1.55	\\
    ELU \cite{elu}	&	\textit{61.97}	$\pm$	0.24	&	51.35	$\pm$	3.01	&	72.57	$\pm$	1.76	&	71.41	$\pm$	1.63	&	\textbf{71.27}	$\pm$	0.58	&	72.06	$\pm$	1.93	\\
    SELU \cite{selu}	&	59.62	$\pm$	0.39	&	64.55	$\pm$	0.43	&	71.47	$\pm$	1.39	&	69.94	$\pm$	1.92	&	55.01	$\pm$	0.98	&	70.15	$\pm$	1.04	\\
    GELU \cite{gelu}	&	61.20	$\pm$	0.61	&	67.25	$\pm$	0.38	&	\textit{74.27}	$\pm$	0.70	&	71.58	$\pm$	0.87	&	71.14	$\pm$	0.29	&	73.31	$\pm$	1.70	\\
    CELU \cite{celu}	&	61.90	$\pm$	0.21	&	55.78	$\pm$	0.69	&	72.87	$\pm$	1.52	&	70.95	$\pm$	1.40	&	63.43	$\pm$	0.81	&	72.68	$\pm$	1.16	\\
    Softplus \cite{softplus}	&	\textbf{62.59}	$\pm$	0.21	&	67.70	$\pm$	0.19	&	73.08	$\pm$	1.66	&	71.99	$\pm$	2.03	&	\textit{71.16}	$\pm$	0.46	&	72.54	$\pm$	1.73	\\
    Swish \cite{swish}	&	59.40	$\pm$	0.41	&	66.05	$\pm$	0.82	&	71.56	$\pm$	1.66	&	71.12	$\pm$	2.08	&	68.42	$\pm$	1.62	&	71.34	$\pm$	1.10	\\
    ABReLU \cite{abrelu}	&	56.21	$\pm$	0.53	&	66.95	$\pm$	0.09	&	71.83	$\pm$	2.26	&	71.96	$\pm$	1.43	&	70.47	$\pm$	0.91	&	\textbf{73.79}	$\pm$	1.45	\\
    LiSHT \cite{lisht}	&	54.09	$\pm$	1.54	&	58.87	$\pm$	0.81	&	66.66	$\pm$	2.50	&	65.28	$\pm$	1.33	&	66.01	$\pm$	1.04	&	65.61	$\pm$	1.10	\\
    SRS \cite{srs}	&	54.93	$\pm$	0.80	&	58.22	$\pm$	1.09	&	70.39	$\pm$	1.09	&	67.11	$\pm$	1.46	&	36.95	$\pm$	0.93	&	64.52	$\pm$	1.39	\\
    Mish \cite{mish}	&	61.81	$\pm$	0.54	&	\textbf{68.13}	$\pm$	0.40	&	73.76	$\pm$	1.48	&	71.89	$\pm$	1.12	&	70.80	$\pm$	0.68	&	73.49	$\pm$	1.39	\\
    PAU \cite{pau}	&	59.81	$\pm$	0.61	&	64.14	$\pm$	0.62	&	70.48	$\pm$	1.53	&	68.59	$\pm$	2.15	&	68.29	$\pm$	0.77	&	67.83	$\pm$	0.35	\\
    PDELU \cite{pdelu}	&	61.35	$\pm$	0.56	&	\textit{67.92}	$\pm$	0.32	&	\textbf{74.48}	$\pm$	1.23	&	\textit{72.11}	$\pm$	1.60	&	70.81	$\pm$	0.47	&	\textit{73.71}	$\pm$	1.64	\\\hline
    \end{tabular}
    }
    \label{tab:results_cifar100}
\end{table*}

\begin{figure*}[!t]
\centering
\begin{subfigure}{.33\textwidth}
\includegraphics[trim=0 0 0 0, width=\columnwidth, clip]{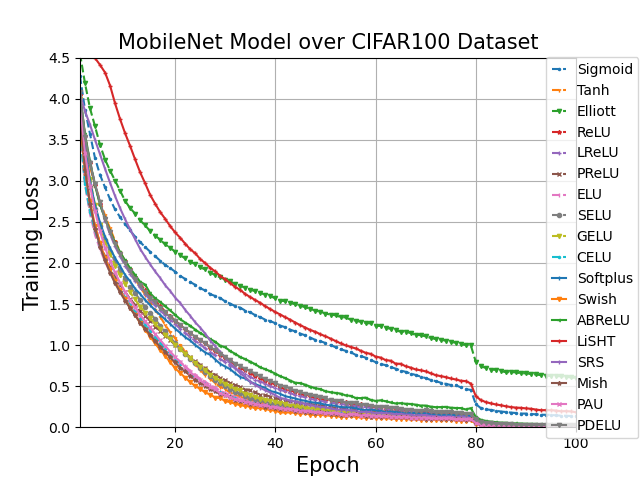}%<left> <lower> <right> <upper>
\end{subfigure}
\begin{subfigure}{.33\textwidth}
\includegraphics[trim=0 0 0 0, width=\columnwidth, clip]{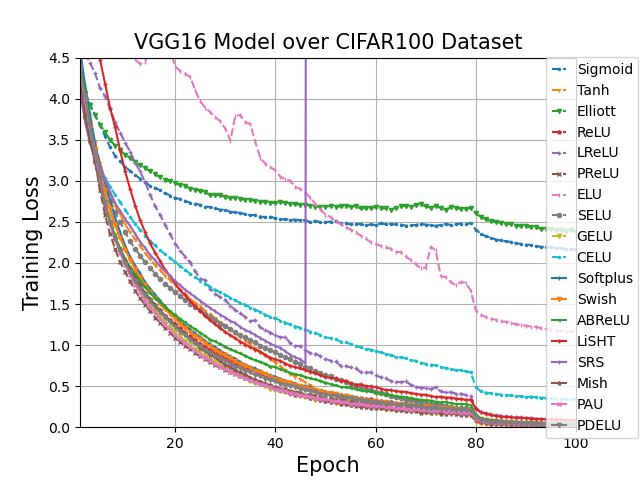}%<left> <lower> <right> <upper>
\end{subfigure}
\begin{subfigure}{.33\textwidth}
\includegraphics[trim=0 0 0 0, width=\columnwidth, clip]{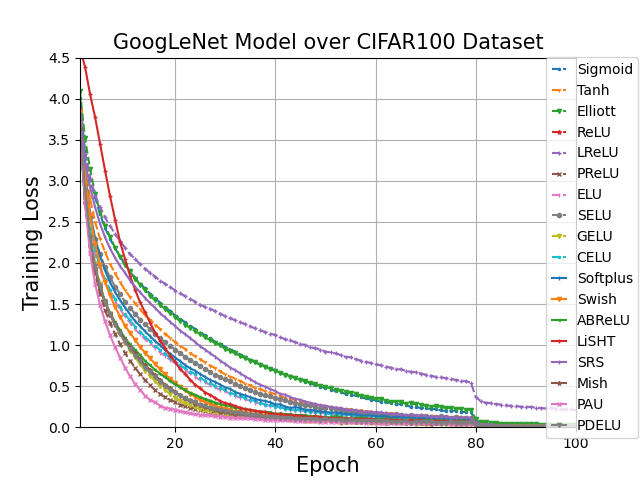}%<left> <lower> <right> <upper>
\end{subfigure}
\begin{subfigure}{.33\textwidth}
\includegraphics[trim=0 0 0 0, width=\columnwidth, clip]{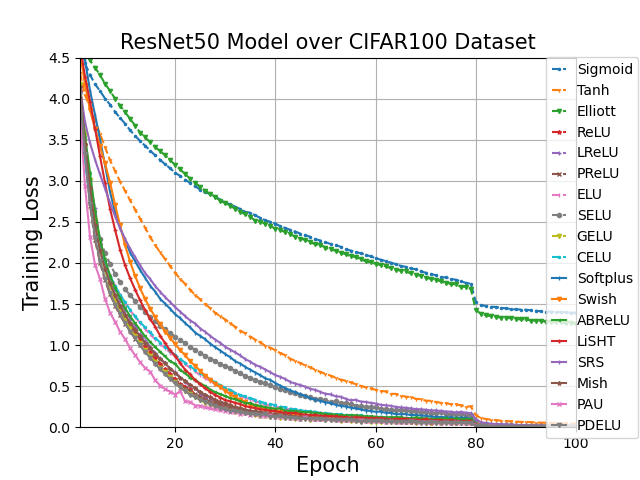}%<left> <lower> <right> <upper>
\end{subfigure}
\begin{subfigure}{.33\textwidth}
\includegraphics[trim=0 0 0 0, width=\columnwidth, clip]{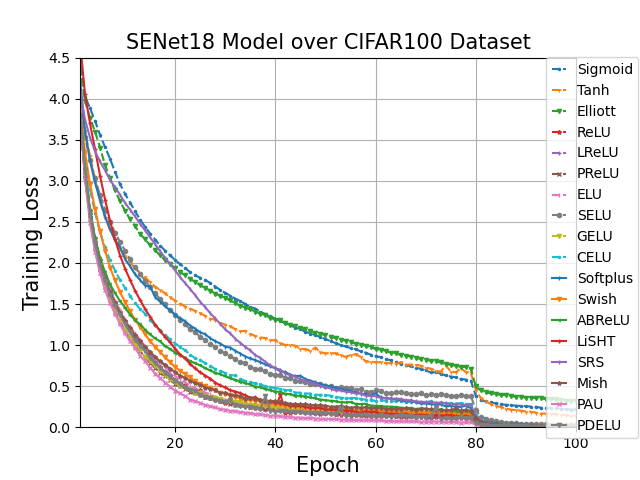}%<left> <lower> <right> <upper>
\end{subfigure}
\begin{subfigure}{.33\textwidth}
\includegraphics[trim=0 0 0 0, width=\columnwidth, clip]{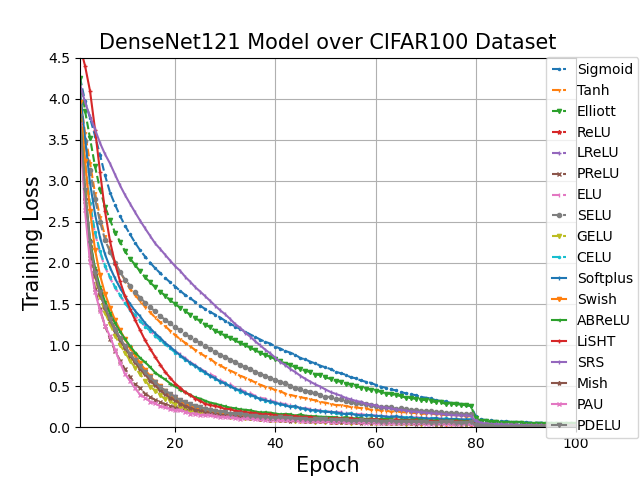}%<left> <lower> <right> <upper>
\end{subfigure}
\caption{Convergence plots over CIFAR100 dataset. }
\label{fig:convergence_analysis}
\end{figure*}

\subsection{Experimental Performance Analysis}
In order to compare the AFs, three experiments are conducted in this paper, including image classification, language translation and speech recognition. Eighteen state-of-the-art AFs are considered for analysis, including Logistic Sigmoid, Tanh, Elliott \cite{farzad2019comparative}, ReLU \cite{alexnet}, LReLU \cite{lrelu} PReLU \cite{prelu}, ELU \cite{elu}, SELU \cite{selu}, GELU \cite{gelu}, CELU \cite{celu}, Softplus \cite{softplus}, Swish \cite{swish}, ABReLU \cite{abrelu}, LiSHT \cite{lisht}, Soft-Root-Sign (SRS) \cite{srs}, Mish \cite{mish}, PAU \cite{pau} and PDELU \cite{pdelu}. Note that Swish, ABReLU, LiSHT, SRS, Mish, PAU and PDELU are the most recent functions. 
Google Colab based computational resource is used in most of the experiments. Few experiments are also performed over a desktop system consisting of 8 GB GPU. The PyTorch framework is used in all the experiments.

The CIFAR10 and CIFAR100 datasets\footnote{\url{https://www.cs.toronto.edu/~kriz/cifar.html}} \cite{cifar} are used for the image classification experiment in this paper. The CIFAR10 dataset contains $50,000$ training images and $10,000$ test images from $10$ object categories. The CIFAR100 dataset contains $50,000$ training images and $10,000$ test images from $100$ object categories. We also utilize the language translation and speech recognition datasets for the experiments. For the experiments over CIFAR-10 and CIFAR-100 datasets, training is performed for 100 Epochs. The batch size is 128 for CIFAR-10 and 64 for CIFAR-100. The learning rate is 0.001 for first 80 Epochs and 0.0001 for last 20 Epochs. Random crop and random horizontal flip are the data augmentation used during training. Data normalization is performed both during train and test times. Adam optimizer is used for the training with cross entropy loss. All existing activation functions except softmax are replaced with the corresponding activation function in different networks.

The test accuracy is reported in Tables \ref{tab:results_cifar} and \ref{tab:results_cifar100} on CIFAR10 and CIFAR100 datasets, respectively.
In these Tables, the mean and standard deviation of image classification accuracy over 5 trials are reported for each AF. Moreover, the better results are highlighted.
Different types of CNN models are used in this experiment, such as plain models (i.e., MobileNet \cite{mobilenet} and VGG16 \cite{vggnet}), inception model (i.e., GoogLeNet \cite{googlenet}) and skip/residual connection based models (i.e., ResNet50 \cite{resnet}, SENet18 \cite{senet}, and DenseNet121 \cite{densenet}). The MobileNet, GoogLeNet and SENet18 are light models, whereas the VGG16, ResNet50 and DenseNet121 are heavy models in terms of the number of trainable parameters. 
Overall, it is observed that the Softplus, ELU and CELU are better suited with MobileNet. The ReLU, Mish and PDELU exhibit good performance with VGG16, GoogleNet and DenseNet. The ReLU, LReLU, ELU, GELU, CELU, ABReLU, and PDELU activation functions are better for the networks having residual connections, such as ResNet50, SENet18 and DenseNet121.
In order to demonstrate the convergence of different AFs, the training loss vs epochs is plotted in Fig. \ref{fig:convergence_analysis} on CIFAR100 dataset using different models. The PAU has emerged as a promising AF with fastest convergence in most of the cases. The PReLU, GELU and PDELU AFs are also consistent with good convergence. Note that the training diverges with SRS for the SENet18 model. Sigmoid and Elliott AFs showed the poorest convergence. 
The time taken for the training is also computed for different AFs using different CNN models on CIFAR100 dataset and reported in Table \ref{tab:results_cifar100_time}. These results are computed using a desktop computer system having 32 GB RAM and 8 GB Nvidia GPU Card for 100 epochs of training. The time is represented in hh:mm:ss format. It is clear that PDELU AF is very inefficient. Moreover, SRS and Elliott also take more time for training. The activations such as ReLU, ELU, CELU, and Softplus depict a good tradeoff between the accuracy and training time.

\begin{table*}[!t]
    \caption{Training time (hh:mm:ss) comparison over CIFAR100 dataset. }
    \centering
    {\footnotesize
    \begin{tabular}{p{0.23\columnwidth}p{0.15\columnwidth}p{0.15\columnwidth}p{0.15\columnwidth}p{0.15\columnwidth}p{0.15\columnwidth}p{0.17\columnwidth}}
    \hline
    Training Time & \multicolumn{6}{c}{CNN Models} \\\cline{2-7}
    Activations & MobileNet & VGG16 & GoogleNet & ResNet50 & SENet18 & DenseNet121 \\\hline
    Sigmoid	&	00:33:15	&	00:49:16	&	04:55:54	&	03:36:03	&	01:13:14	&	04:12:24	\\
    Tanh	&	00:33:18	&	00:49:55	&	04:58:02	&	03:33:03	&	01:13:18	&	04:09:24	\\
    Elliott  \cite{farzad2019comparative}	&	00:49:52	&	00:59:13	&	06:53:55	&	05:38:49	&	01:41:38	&	07:46:55	\\
    ReLU \cite{alexnet}	&	00:31:22	&	00:47:19	&	04:55:10	&	03:32:30	&	01:15:33	&	04:15:06	\\
    LReLU \cite{lrelu}	&	00:31:48	&	00:49:03	&	05:01:30	&	03:33:00	&	01:18:38	&	04:14:09	\\
    PReLU \cite{prelu}	&	00:44:24	&	00:49:01	&	05:42:18	&	03:55:57	&	01:27:05	&	04:55:47	\\
    ELU \cite{elu}	&	00:31:05	&	00:47:38	&	04:57:37	&	03:36:47	&	01:13:25	&	04:08:39	\\
    SELU \cite{selu}	&	00:29:40	&	00:47:31	&	04:54:57	&	03:33:47	&	01:13:27	&	04:09:17	\\
    GELU \cite{gelu}	&	00:29:43	&	00:47:22	&	04:55:53	&	03:32:32	&	01:13:32	&	04:11:26	\\
    CELU \cite{celu}	&	00:29:36	&	00:46:47	&	05:00:44	&	03:31:40	&	01:14:08	&	04:18:11	\\
    Softplus \cite{softplus}	&	00:29:44	&	00:47:06	&	04:58:55	&	03:32:03	&	01:14:02	&	04:12:08	\\
    Swish \cite{swish}	&	00:43:13	&	00:55:37	&	06:18:38	&	04:58:38	&	01:32:15	&	06:41:14	\\
    ABReLU \cite{abrelu}	&	00:38:51	&	00:53:49	&	05:43:59	&	04:27:02	&	01:25:30	&	05:42:53	\\
    LiSHT \cite{lisht}	&	00:37:01	&	00:54:10	&	05:40:00	&	04:25:57	&	01:23:59	&	05:38:15	\\
    SRS \cite{srs}	&	01:06:38	&	01:11:36	&	08:43:09	&	07:35:35	&	02:05:33	& 11:10:27 \\
    Mish \cite{mish}	&	00:40:19	&	00:54:23	&	05:59:48	&	04:46:45	&	01:28:53	&	06:10:27	\\
    PAU \cite{pau}	&	00:41:59	&	00:54:10	&	05:54:22	&	04:12:31	&	01:25:37	&	05:39:57	\\
    PDELU \cite{pdelu}	&	05:23:38	&	04:01:55	&	34:22:00	&	36:48:48	&	08:32:40	&	50:23:00	\\\hline
    \end{tabular}
    }
    \label{tab:results_cifar100_time}
\end{table*}

\begin{table}[!t]
    \caption{Experimental results for German to English language translation and speech recognition tasks.}
    \centering
    {\footnotesize
    \begin{tabular}{p{0.23\columnwidth}p{0.2\columnwidth}p{0.01\columnwidth}p{0.15\columnwidth}p{0.15\columnwidth}}
    \hline
    \centering
     & Language Translation && \multicolumn{2}{c}{Speech Recognition} \\\cline{2-2}\cline{4-5}
    Activations & Bleu Score && Average CER & Average WER \\\hline
    Sigmoid & 14.59	$\pm$	0.47 &	&	0.53	$\pm$	0.18	&	1.19	$\pm$	0.39\\
    Tanh & \textbf{20.93}	$\pm$	0.91 &	&	0.26	$\pm$	0	&	0.68	$\pm$	0\\
    Elliott  \cite{farzad2019comparative} & 14.49	$\pm$	0.96 &	&	0.40	$\pm$	0.01	&	0.93	$\pm$	0.01\\
    ReLU \cite{alexnet} & 18.88	$\pm$	0.86 &	&	\textbf{0.24}	$\pm$	0.01	&	\textit{0.66}	$\pm$	0.01\\
    LReLU \cite{lrelu} & 18.89	$\pm$	0.82 &	&	\textbf{0.24}	$\pm$	0	&	\textit{0.66}	$\pm$	0.01\\
    PReLU \cite{prelu} & 20.04	$\pm$	0.98 &	&	\textbf{0.24}	$\pm$	0	&	\textbf{0.65}	$\pm$	0\\
    ELU \cite{elu} & 19.40	$\pm$	1.33 &	&	\textit{0.25}	$\pm$	0	&	0.67	$\pm$	0\\
    SELU \cite{selu} & \textit{20.85}	$\pm$	0.64 &	&	0.26	$\pm$	0	&	0.69	$\pm$	0.01\\
    GELU \cite{gelu} & 18.75	$\pm$	1.83 &	&	\textbf{0.24}	$\pm$	0	&	\textbf{0.65}	$\pm$	0\\
    CELU \cite{celu} & 18.71	$\pm$	0.55 &	&	\textit{0.25}	$\pm$	0	&	0.67	$\pm$	0\\
    Softplus \cite{softplus} & 16.78	$\pm$	0.84 &	&	0.30	$\pm$	0.01	&	0.76	$\pm$	0.02\\
    Swish \cite{swish} & 19.51	$\pm$	0.97 &	&	\textbf{0.24}	$\pm$	0.01	&	\textbf{0.65}	$\pm$	0.01\\
    ABReLU \cite{abrelu} & 17.55	$\pm$	0.63 &	&	\textit{0.25}	$\pm$	0	&	0.68	$\pm$	0\\
    LiSHT \cite{lisht} & 20.39	$\pm$	0.93 &	&	0.29	$\pm$	0.01	&	0.74	$\pm$	0.01 \\
    SRS \cite{srs} & 20.66	$\pm$	0.78 &	&	0.28	$\pm$	0	&	0.72	$\pm$	0 \\
    Mish \cite{mish} & 19.56	$\pm$	1.15 &	&	\textbf{0.24}	$\pm$	0	&	\textbf{0.65}	$\pm$	0 \\
    PAU \cite{pau} & 20.11	$\pm$	1.24 &	&	\textbf{0.24}	$\pm$	0	&	\textbf{0.65}	$\pm$	0.01 \\
    PDELU \cite{pdelu} & 19.07	$\pm$	0.95 &	&	\textit{0.25}	$\pm$	0	&	0.67	$\pm$	0.01\\\hline
    \end{tabular}
    }
    \label{tab:results_language_speech}
\end{table}

The results for language translation and speech recognition for different AFs are illustrated in Table \ref{tab:results_language_speech}. The German to English translation is used to test the performance of the AFs over text data. Benchmark Seq2Seq model consisting of a Long Short Term Memory (LSTM) based autoencoder network is used for the experiment. The model and dataset are downloaded from Kaggle\footnote{https://www.kaggle.com/parthplc/pytorch-seq2seq-machine-translation/notebook}. The AF is applied to the feature embedding before the dropout layer. 
For the language translation experiments, the number of Epochs is set to 50 with 0.001 learning rate and 256 batch size. The embedding size of encoder and decoder is 300. 
% The hidden size is 1024 with 1 layer. 
The dropout factor is 0.5 for both encoder and decoder. Adam optimizer is used for the training with cross entropy loss.
The Bleu score \cite{bleu} with $4$\textit{-gram} is reported in Table \ref{tab:results_language_speech} in $2^{nd}$ column for different AFs. The mean and standard deviation of Bleu score over 5 trials are reported for each AF. It is noticed that the Tanh and SELU AFs are better suitable for language translation. The PReLU, LiSHT, SRS and PAU AFs also perform better for language translation.

The speech recognition experiment is also performed to show the performance of the different AFs for time-series signal data. The end-to-end speech recognition based Deep Speech 2 framework available from assemblyai\footnote{https://www.assemblyai.com/blog/end-to-end-speech-recognition-pytorch} is used. The model consists of $2$ layers of residual convolution layers to learn the relevant audio features, and $2$ layers of bidirectional gated recurrent units (GRUs) to use the learned residual convolutional audio features. The $100$ hours of transcribed audio English data from LibriSpeech dataset is used for the experiment. 
For the speech recognition experiments, torchaudio 0.4.0 and torch 1.4.0 are used. The model consists of 2 CNN layers and 2 RNN layers. The dimension of a RNN layer is 512. Number of classes is 29 in the dataset. Dropout factor is 0.5. The learning rate is 0.0005, batch size is 10 and the number of Epochs is 10.
The mean and standard deviation over 5 trials of character error rate (CER) and word error rate (WER) are reported in Table \ref{tab:results_language_speech} for speech recognition. The recent AFs such as PReLU, GELU, Swish, Mish and PAU AFs are found as the most suitable for speech recognition in this experiment.

\section{Conclusion and Recommendations}
\label{conclusion}
An extensive and up to date survey of activation functions is conducted in this paper. Different types of AFs are considered, including Logistic Sigmoid and Tanh based, ReLU based, ELU based, and Learning based. However, the main focus is given to the recent developments in AFs in view of the deep learning applications of neural networks. 
The overview of AFs presented in this paper focuses on the aspects including the detailed coverage of AFs, classification and performance comparison over image, text and speech data.

Following are the concluding remarks of the survey and performance analysis conducted through this paper:
\begin{itemize}
    \item Most of the improvements in Logistic Sigmoid and Tanh targets to tackle the non zero-mean and zero-gradient problems. However, these improvements carry forward the drawback of increased complexity.
    \item The ReLU variants try to tackle the three major problems of ReLU, namely under-utilization of negative values, limited nonlinearity and unbounded output. 
    These activations perform well for some applications, e.g. LReLU and ABReLU works better with residual networks. However, most of these activations fail to perform better than ReLU, e.g. LReLU, PReLU and ABReLU do not improve for MobileNet, VGG and GoogleNet models.
    Note that, the ReLU, Leaky ReLU and PReLU AFs are the most common choice among researchers due to its simplicity. Moreover, many networks consider the ReLU as a default choice for the AF.
    \item The exponential based AFs also focus over the better utilization of the negative values and to avoid the saturation for important features. However, most of the exponential activations suffer due to the 
    % increased computational complexity and 
    non-smooth functions.
    \item The learning based adaptive AFs try to find the best parameters to represent the non-linearity needed for the given dataset. This category of AF has gained more popularity in recent years. However, the major problem associated with such AF is to find the better base function and number of trainable parameters. Some AFs diverge during the training if not initialized properly.
    \item In contrast to existing surveys, this survey covers an exhaustive list different types of AFs. Moreover, a performance analysis on different types of data using several AFs provides new insights for future research.
\end{itemize}

Following are the recommendations curated from this survey and performance analysis:
\begin{itemize}
    \item In order to speed up the training, both negative \& positive values should be used to ensure the near zero mean.
    \item The most important aspect in deep learning is to find the network having matching complexity as the dataset complexity. If the complexity of the model is high then it may lead to overfitting and if the complexity of the model is low then it may lead to under convergence. Thus, the AF should bridge this gap based on the model and dataset complexity during training automatically.
    \item The Logistic Sigmoid and Tanh AFs should be avoided for Convolutional Neural Networks as it leads to poor convergence. However, this type of AF is commonly used as gates in recurrent neural networks.
    \item Despite the ReLU being a popular choice, recently proposed AFs such as Swish, Mish, and PAU are also worth trying for different problems.
    \item The ReLU, Mish and PDELU activation functions have shown a good performance with VGG16 and GoogleNet. The ReLU, LReLU, ELU, GELU, CELU, and PDELU functions are better for the networks having residual connections for image classification.
    \item In general, the parametric AFs show better convergence as it can adapt the data faster by learning the parameter from the data. Specially, PAU, PReLU and PDELU have shown better convergence.
    \item Some AFs lead to increased training time complexity. PDELU and SRS are such examples. However, AFs such as ReLU, SELU, GELU, and Softplus depict a promising tradeoff between the accuracy and training time.
    \item The exponential AFs generally lead to the increased non-linearity due to utilization of the negative values.
    \item The Tanh and SELU AFs are found better for language translation along with PReLU, LiSHT, SRS and PAU. 
    \item It is suggested to use the PReLU, GELU, Swish, Mish and PAU AFs for speech recognition.
\end{itemize}

\bibliographystyle{elsarticle-num}
\bibliography{References}

\end{document}